\documentclass[dvipsnames,pbalance,format=sigconf,anonymous=false,review=false,nonacm=true]{acmart}
\usepackage{amsmath}
\usepackage[dvipsnames,table]{xcolor}
\usepackage{graphicx} 
\usepackage{tabularx}
\usepackage{longtable}
\usepackage{booktabs}
\usepackage{multicol}
\usepackage[inline]{enumitem}
\usepackage{algorithm}
\usepackage{algorithmicx}
\usepackage{algpseudocode}
\usepackage{subcaption}
\usepackage[most]{tcolorbox}
\usepackage{placeins,soul}
\usepackage{pgfplotstable}
\usepackage{pgffor}
\usetikzlibrary{calc}
\usepackage{glossaries}
\usepackage{hyperref}
\usepackage{orcidlink}
\usepackage{bookmark}
\usepackage[nameinlink]{cleveref}

\usepackage{bibunits}
\usepackage{bibentry}

\makeatletter
\newcommand{\nobibliographybbl}[1]{%
  \IfFileExists{#1.bbl}{%
    \begingroup
      \renewenvironment{thebibliography}[1]{%
        \usecounter{enumiv}%
        \renewcommand\item[1][]{%
          \ifx\relax####1\relax\stepcounter\@listctr\fi}%
        \let\BR@bibitem=\bibitem
        \let\bibitem=\BR@b@bibitem
      }{}%
      \let\@biblabel\@gobble
      \input{#1.bbl}%
    \endgroup
  }{%
  }%
}
\makeatother

\defaultbibliographystyle{ACM-Reference-Format}
\defaultbibliography{biblio}

\newcolumntype{C}[1]{>{\centering\arraybackslash}m{#1}}

\tcbset {
  base/.style={
    arc=0mm, 
    bottomtitle=0.5mm,
    boxrule=0mm,
    colbacktitle=black!10!white, 
    coltitle=black, 
    fonttitle=\bfseries, 
    left=2.5mm,
    leftrule=1mm,
    right=3.5mm,
    title={#1},
    toptitle=0.75mm, 
    breakable
  }
}

\definecolor{brandblue}{rgb}{0.34, 0.7, 1}
\newtcolorbox{mainbox}[1]{
  colframe=brandblue, 
  base={#1}
}

\newtcolorbox{subbox}[1]{
  colframe=black!30!white,
  base={#1}
}

\hypersetup{
  colorlinks=true, 
  urlcolor=black, 
  linkcolor=black, 
  citecolor=black 
}

\newacronym{aaai}{AAAI}{Association for the Advancement of Artificial Intelligence}
\newacronym{acm}{ACM}{Association for Computing Machinery}
\newacronym{ai}{AI}{Artificial Intelligence}
\newacronym{bbsr}{BBSR}{Benchmarking, Benchmarks, Software and Reproducibility}
\newacronym{ec}{EC}{Evolutionary Computation}
\newacronym{gecco}{GECCO}{Genetic and Evolutionary Computation Conference}
\newacronym{llm}{LLM}{Large Language Model}
\newacronym{ml}{ML}{Machine Learning}
\newacronym{neurips}{NeurIPS}{Neural Information Processing Systems}
\newacronym{aad}{AAD}{Automated Algorithm Design}
\newacronym{ast}{AST}{Abstract Syntax Tree}

\setcopyright{acmlicensed}
\acmDOI{10.1145/nnnnnnn.nnnnnnn} 
\acmISBN{978-x-xxxx-xxxx-x/YY/MM} 
\acmConference[GECCO '26]{The Genetic and Evolutionary Computation Conference 2026}{July 13--17, 2026}{San José, Costa Rica}
\acmYear{2026}
\copyrightyear{2026}

\begin{document}
\begin{bibunit}
\title{Structural bias in multi-objective optimisation}
%
%
%
\author{Jakub Kudela\orcidlink{0000-0002-4372-2105}}
\email{jakub.kudela@vutbr.cz}
\orcid{0000-0002-4372-2105}
\affiliation{%
  \institution{Brno University of Technology}
  \country{Czech Republic}
}

\author{Niki van Stein\orcidlink{0000-0002-0013-7969}}
\email{n.van.stein@liacs.leidenuniv.nl}
\orcid{0000-0002-0013-7969}
\affiliation{%
  \institution{LIACS, Leiden University}
  \country{Netherlands}
}

\author{Thomas B{\"a}ck\orcidlink{0000-0001-6768-1478}}
\email{t.h.w.baeck@liacs.leidenuniv.nl}
\orcid{0000-0001-6768-1478}
\affiliation{%
  \institution{LIACS, Leiden University}
  \country{Netherlands}
}

\author{Anna V. Kononova\orcidlink{0000-0002-4138-7024}}
\email{a.kononova@liacs.leidenuniv.nl}
\orcid{0000-0002-4138-7024}
\affiliation{%
  \institution{LIACS, Leiden University}
  \country{Netherlands}
}

\renewcommand{\shortauthors}{Kudela et al.}

\begin{abstract}
Structural bias (SB) refers to systematic preferences of an optimisation algorithm for particular regions of the search space that arise independently of the objective function. While SB has been studied extensively in single-objective optimisation, its role in multi-objective optimisation remains largely unexplored. This is problematic, as dominance relations, diversity preservation and Pareto-based selection mechanisms may introduce or amplify structural effects.

In this paper, we extend the concept of structural bias to the multi-objective setting and propose a methodology to study it in isolation from fitness-driven guidance. We introduce a suite of synthetic multi-objective test problems with analytically controlled Pareto fronts and deliberately uninformative objective values. These problems are designed to decouple algorithmic behaviour from problem structure, allowing bias induced purely by algorithmic operators and design choices to be observed. The test suite covers a range of Pareto front shapes, densities and noise levels, enabling systematic analysis of different manifestations of structural bias.

We discuss methodological challenges specific to the multi-\\objective case and outline how existing SB detection approaches can be adapted. This work provides a first step towards behaviour-based benchmarking of multi-objective optimisers, complementing performance-based evaluation and informing more robust algorithm design.
\end{abstract}

\ccsdesc[500]{Theory of computation~Continuous optimization}
\ccsdesc[300]{Mathematics of computing~Continuous optimization}
\ccsdesc[500]{Computing methodologies~Randomized search}

\keywords{Multi-objective Optimization, Continuous Optimization, Structural Bias, Algorithm Behaviour, Benchmarking}

\maketitle

\glsresetall

\section{Introduction}
The growing diversity and complexity of black-box optimisation problems have led to a large ecosystem of heuristic optimisation algorithms. Since no single algorithm performs best across all problem classes\footnote{with some caveats in the case of multiobjective optimisation~\cite{Corne2003_Some,Corne2003_No}}~\cite{NFLT}, systematic benchmarking is essential to understand when and why particular methods succeed or fail~\cite{bart20gArxiv}. 
Most established benchmarks for continuous single- and multi-objective optimisation are performance-driven, such as the Black-Box Optimisation Benchmark (BBOB)~\cite{hansen2021coco} for single-objective optimisation (SOO) and ZDT~\cite{ZDT}, DTLZ~\cite{DTLZ}, WGF~\cite{WGF} and bbob-biobj~\cite{Brockhoff2022_Using} suites for multi-objective optimisation (MOO). They mainly focus on performance metrics like convergence speed or final objective value for single-objective and Hypervolume and Inverted Generational Distance for multi-objective benchmarks. While indispensable, such benchmarks provide limited insight into \emph{how} algorithms behave internally during the search process.

Behaviour-based benchmarking complements performance-based evaluation by analysing algorithmic dynamics independently of problem-specific fitness guidance. One important behavioural property is \emph{structural bias} (SB)~\cite{Kononova2015}: systematic preferences of an algorithm for particular regions of the search space that arise solely from algorithmic design choices, rather than the objective function. Structural bias is often undesirable in general-purpose optimisers, as it may prevent algorithms from locating optima in certain regions of the domain, even when sampled information on the objective function values does not justify such preferences.

To study SB in isolation for single-objective black-box optimisation, the BIAS toolbox~\cite{bib:BIAS} was introduced as a principled methodology based on repeated optimisation runs on a deliberately uninformative random objective function $f_0$ introduced in \cite{Kononova2015}. Deviations from uniformity in the resulting final-point distributions indicate the presence and type of structural bias. This approach has enabled large-scale, automated SB detection and has revealed that many state-of-the-art single-objective optimisers exhibit non-trivial forms of bias~\cite{bib:BIAS,deepbias,van2024_deepdive,vStein2021_emergence,Kononova2020CEC}.

Structural bias has so far been studied exclusively in the single-objective setting. This is a significant gap: multi-objective optimisation introduces dominance relations, diversity preservation mechanisms and Pareto-based selection that may introduce new forms of bias or amplify existing ones. Understanding whether, how and where multi-objective optimisers exhibit structural bias is therefore essential for reliable behaviour-based benchmarking in this setting.

In this work, we extend the concept of structural bias to multi-objective black-box optimisation. We propose a methodology that decouples algorithmic behaviour from fitness-driven guidance by means of synthetic multi-objective test problems with analytically controlled Pareto fronts and deliberately uninformative objective values. This allows structural bias induced purely by algorithmic operators and selection mechanisms to be observed and analysed in isolation. Our study provides a first step towards behaviour-based benchmarking of multi-objective optimisers and complements existing performance-oriented evaluation practices.

The remainder of this paper is structured as follows. Section~\ref{sect:SB} revisits structural bias and existing detection methodologies for single objective optimisation. Section~\ref{sect:mo_sb} introduces the proposed multi-objective test problems and experimental setup. Section~\ref{sect:discuss} contains a discussion of results, while Section~\ref{sect:conclusions} presents conclusions and outlook.

\section{Structural bias}\label{sect:SB}
Optimisation algorithms are typically designed to be generic, targeting wider classes of problems rather than individual instances. While we would like such algorithms to perform well across different problems, the No-Free Lunch Theorem states that no optimiser can be optimal over all possible optimisation problems~\cite{NFLT}. In practice, algorithms therefore tend to specialise: they perform well on certain problem classes and poorly on others. Identifying which problems suit which algorithms is non-trivial and has motivated extensive research into landscape features and instance-space analysis~\cite{SMITHMILES201412,mersmann2011exploratory}. 

One property that should \emph{not} characterise a useful class of problems is the location of the optimum or, in case of multi-objective optimisation, the location of the Pareto optimal set~\cite{Deb2001} within the search domain. Regardless of the number of objectives, an algorithm that consistently succeeds only when optima lie near specific regions, such as the centre, the origin or the domain boundaries, is of limited practical value. Ideally, an optimisation algorithm should be able to locate optima regardless of their position in the search space and should ideally do so with comparable effort across different optima locations.

In iterative optimisation algorithms, candidate solutions are generated and transformed through repeated application of algorithmic operators. Progress towards an optimum is intended to be driven solely by information obtained from the objective function(s), for example, through fitness differences or dominance relations. However, the iterative structure of the algorithm itself may introduce additional influences on the search dynamics. When, over multiple runs, these influences induce systematic preferences for certain regions of the search space that are independent of the sampled values of the objective function(s), we refer to this phenomenon as \emph{structural bias} (both in single- and multi-objective settings).

Structural bias becomes particularly relevant in the multi-objective optimisation, as dominance relations, diversity preservation mechanisms and Pareto-based selection may introduce additional location-dependent effects that are absent or less pronounced in the single-objective setting.

\subsection{Existing methodology for measuring SB in the single-objective setting}\label{sect:}
Detecting structural bias is challenging because algorithmic dynamics and fitness landscape information are intrinsically coupled. To separate these effects, a \textit{dedicated test function}, denoted \(f_0\), was introduced in~\cite{Kononova2015}. The function eliminates informative fitness feedback by assigning independent random values to all points in the domain. As a result, any systematic non-uniformity observed in the distribution of the best solutions across repeated runs can be attributed solely to algorithm-induced effects. The function \(f_0\) is defined as\footnote{Throughout this paper, one-dimensional uniform sampling within $[a,b]$ is denoted by $\mathcal{U}(a,b)$, and sampling from a normal distribution with mean $\mu$ and standard deviation $\sigma$ by $\mathcal{N}(\mu,\sigma)$.}
\begin{equation}
    f_0 : [0,1]^d \rightarrow [0,1], \quad \text{where } \forall x,\; f_0(x) \sim \mathcal{U}(0,1).
\end{equation}

To inspect an algorithm for structural bias, it is run on the function \(f_0\) for a statistically meaningful number of independent runs, using a fixed dimensionality and an evaluation budget representative of practical use. The locations of the best solution obtained in each run are then analysed for uniformity. In early SB studies, this analysis relied exclusively on visual inspection, typically by displaying the best found solutions component-wise using parallel coordinate plots \cite{bib:inselberg1985plane}. Subsequent work investigated a range of statistical tests for uniformity \cite{Kononova2015,Kononova2020CEC,Kononova2020PPSN,vStein2021_emergence}, ultimately leading to the development of the BIAS toolbox. This framework employs an ensemble of statistical tests, complemented by a deep-learning-based classification component, to provide an automated, objective and systematic assessment of SB (see Section~\ref{sec:toolbox}).

Alternative SB detection methodologies include the examination of algorithms on perfectly symmetrical problems with two different local optima \cite{walden2024simple}, or the aggregation of the searched points into one dimension and comparing a discretised distribution of the points with $\mathcal{U}(0,1)$ via the chi-squared test \cite{ibehej2025investigation}.

It is interesting to mention that recently the wider machine learning community has re-invented the concept of structural bias under the name of ``inductive bias'', claiming that the choice of optimiser influences the qualitative properties of the learnt solutions and urging the community to understand the biases of already existing methods, rather than solely judging them based on their convergence rates~\cite{Pascanu2025,Scaife2025}. 

\begin{table*}[!t]
\caption{Proposed parameterised test problems in the form $[g_1,g_2]$ selected for detection of structural bias in the multi-objective setting. The table also includes auxiliary values for construction of $g_i$, corresponding parameter values $r$, reasons for considering each function, Pearson correlation coefficient $\rho$ between $g_i$ computed on a random sample of $10^5$ random points evaluated on each problem (values marked with a * are not sample-based) and estimates on the sizes of Pareto Fronts $\left| PF \right|_{10^4}$ computed from evaluating a sample of $10^4$ random points. Shapes of Pareto fronts are shown in Figure~\ref{fig:problems} and results on the investigation of stability of estimates $\left| PF \right|_{10^4}$ are given in Figure~\ref{fig:proportions}.\vspace{-3mm}}\label{tab:problems}
\resizebox{0.95\textwidth}{!}{
\begin{tabular}{C{1.05cm} C{4.35cm} C{0.8cm} C{4.85cm} C{1.1cm} C{1.45cm} C{0.7cm} C{0.7cm}}
    \toprule
    \textbf{Problem} & \textbf{auxiliary values} & $g_1$ & $g_2$ & $r$ \textbf{value} & \textbf{why?} & $\rho$ & $\left| PF \right|_{10^4}$\\
    \midrule

    $f_1$ &
    -- &
    $\mathcal{U}(0,1)$ &
    $\mathcal{U}(0,1)$ &
    -- & $f_0$ based & $0.00^*$ & $10$\\
    \midrule

    \shortstack{$f_2^\alpha$\\ $f_2^\beta$\\ $f_2^\gamma$} &
    $a=\mathcal{U}(0,1)$ &
    $a$ &
    $1-\sin(12\cdot\pi\cdot a^{1.5})/4-\sqrt{\frac{a}{2}}+r\cdot(\mathcal{N}(0,1))^2$ &
    \shortstack{$2$\\\\\\ $0.08$\\\\\\ $0.0015$} &
    disconnected PF & \shortstack{$-0.03$\\\\\\ $-0.36$\\\\\\ $-0.43$} & \shortstack{$10^2$\\\\ $3\cdot 10^2$\\\\ $10^3$} \\
    \midrule

    \shortstack{$f_3^\alpha$\\ $f_3^\beta$\\ $f_3^\gamma$} &
    \shortstack{
        $a=\mathcal{U}(0,1)$\\
        $p=[0.1,0.2,\dots,0.9]$\\
        $s=[1,0.9,\dots,0.1]$\\
        $b=\sum_{i=1}^9 \mathbf{1}_{a > p(i)}$
    } &
    $a$ &
    $\frac{0.1}{(10\cdot(a-\frac{b}{10})+1)^{10}} + s_{b+1} + r\cdot(\mathcal{N}(0,1))^2$ &
    \shortstack{$20$\\\\\\ $0.9$ \\\\\\ $0.02$} &
    convex stairs & \shortstack{$-0.01$\\\\\\ $-0.22$\\\\\\ $-0.99$} & \shortstack{$10^2$\\\\ $3\cdot10^2$\\\\ $10^3$} \\
    \midrule

    \shortstack{$f_4^\alpha$\\ $f_4^\beta$\\ $f_4^\gamma$} &
    \shortstack{
    $a=\mathcal{U}(0,1)$\\
    $b=\begin{cases}
    -2\cdot a^2, & a < 0.5\\
    -2\cdot (a-0.5)^2 - 0.5, & a \geq 0.5
    \end{cases}$
    } &
    $a$ &
    $b +r\cdot(\mathcal{N}(0,1))^2$ &
    \shortstack{$60$\\\\\\ $2$\\\\\\ $0.06$} &
    concave regions & \shortstack{$0.00$\\\\\\ $-0.01$\\\\\\ $-0.95$} & \shortstack{$10^2$\\\\ $3\cdot10^2$\\\\ $10^3$} \\
    \midrule

    $f_5$ &
    $a=\mathcal{U}(0,1)$ &
    $a$ &
    $-a$ &
    -- &
    correlated & $-1.00^*$ & $10^4$\\
    \bottomrule
\end{tabular}
}
\end{table*}

\subsection{BIAS toolbox}\label{sec:toolbox} 
BIAS~\cite{bib:BIAS} is an open-source Python toolbox for detecting structural bias in single-objective continuous optimisation algorithms. The toolbox operationalises SB detection by analysing the distributions of final solutions obtained from repeated optimisation runs on the random test function $f_0$. Structural bias and its type are inferred through an ensemble of $39$ statistical tests, whose outcomes are aggregated and interpreted using a Random Forest classifier.
More recently, Deep-BIAS~\cite{deepbias} extended this framework by replacing extensive statistical testing with a deep learning model that directly classifies the presence and type of SB from raw performance distributions. This significantly improves detection accuracy, enables fine-grained bias classification and allows explanations to be derived using explainable AI techniques. Deep-BIAS has made it feasible to analyse structural bias across hundreds of algorithm configurations at scale.

Beyond detection, other work has started to investigate the \emph{consequences} of structural bias. In particular, an in-depth analysis of modular CMA-ES variants~\cite{van2024_deepdive} demonstrated that algorithm components inducing specific types of SB can interact with landscape properties and optimum locations, leading to performance differences even on closely related problems. These results underline that structural bias is a factor that can materially affect algorithm robustness.

\section{Structural bias in multi-objective optimisation algorithms}\label{sect:mo_sb}
While one of the fundamental differences between MOO vs SOO is reasoning in the objective space, SB analysis concerns only the decision space~\cite{Kononova2015}. However, studying SB in multi-objective optimisation still requires a substantial \textit{extension} of the methodological setup. While in single-objective optimisation the result of the run is \textit{a single solution} (optimal or the best solution found over the run), in multi-objective optimisation we get \textit{a set of non-dominated points} (Pareto set or its approximation). 

As the first step in the investigation of SB in MOO, in this study, we limit the presented methodology to the case of \textit{two} objectives.

\subsection{Choice of test problems}\label{sect:funcs}
Traditionally, the difficulty of MOO problems is defined (in part) by the shape and size of the True Pareto front~\cite{Deb2001}, which represents a set of compromises between multiple (typically) conflicting objectives. When designing new test functions, the space of possible Pareto front geometries and of proportions between dominated and Pareto-optimal solutions is effectively infinite. As concrete choices are therefore unavoidable, we aimed to select a range of diverse Pareto front shapes and sizes, representing different sources of difficulty in multi-objective optimisation, in order to assess whether and how these characteristics influence the manifestation of structural bias.

Following the reasoning of the SOO case, the proposed test function(s) must be able to decouple algorithmic dynamics from landscape information. Our current best approach for achieving this is the use of random objective functions. However, employing $f_0$ for all objectives is insufficient in the multi-objective setting, as it removes the primary source of difficulty in MOO, namely the presence of conflicting objectives.

\begin{figure*}
    \centering
   \begin{tabular}{ccc}
   \includegraphics[width=0.2\linewidth]{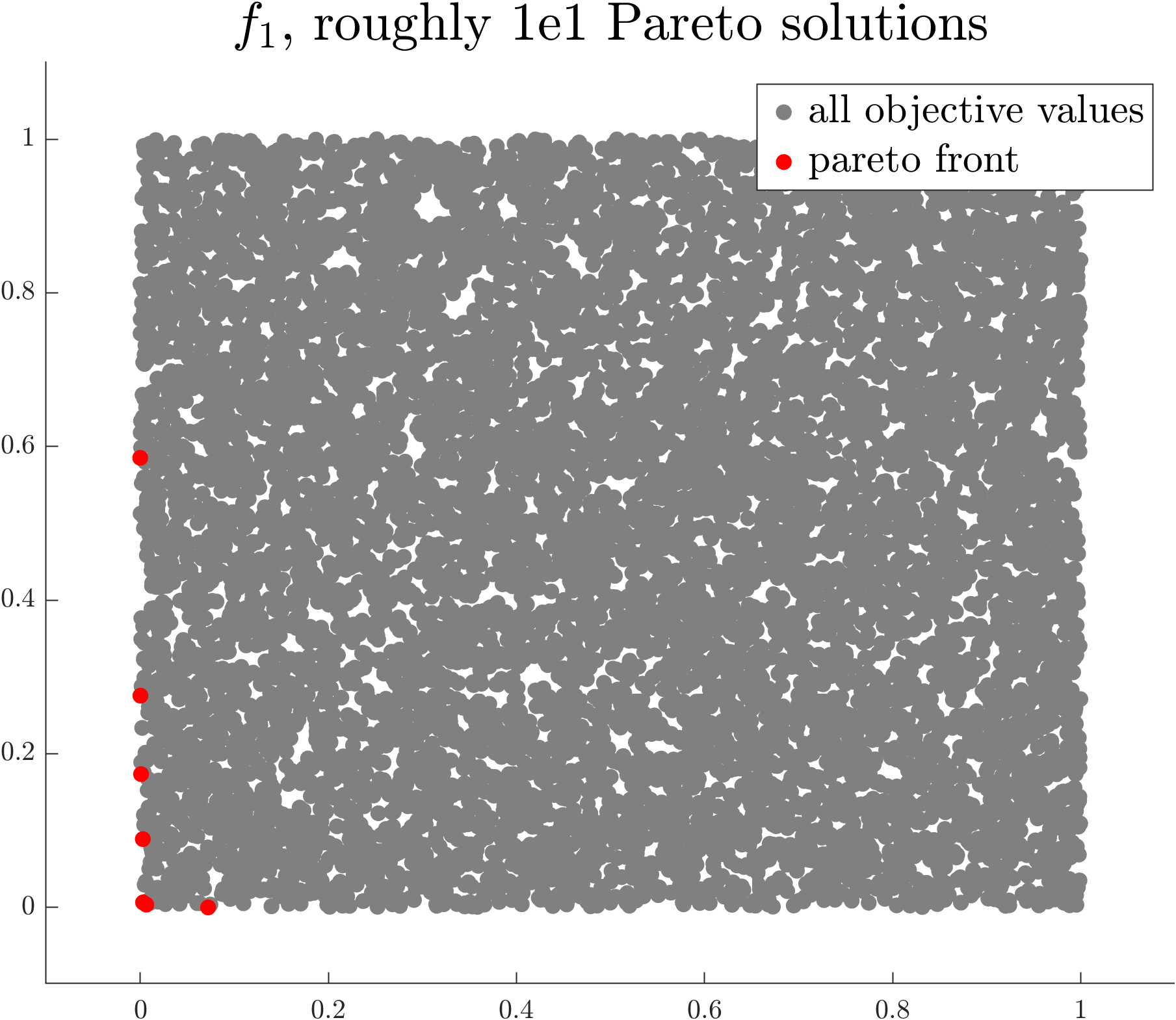}   & \includegraphics[width=0.2\linewidth]{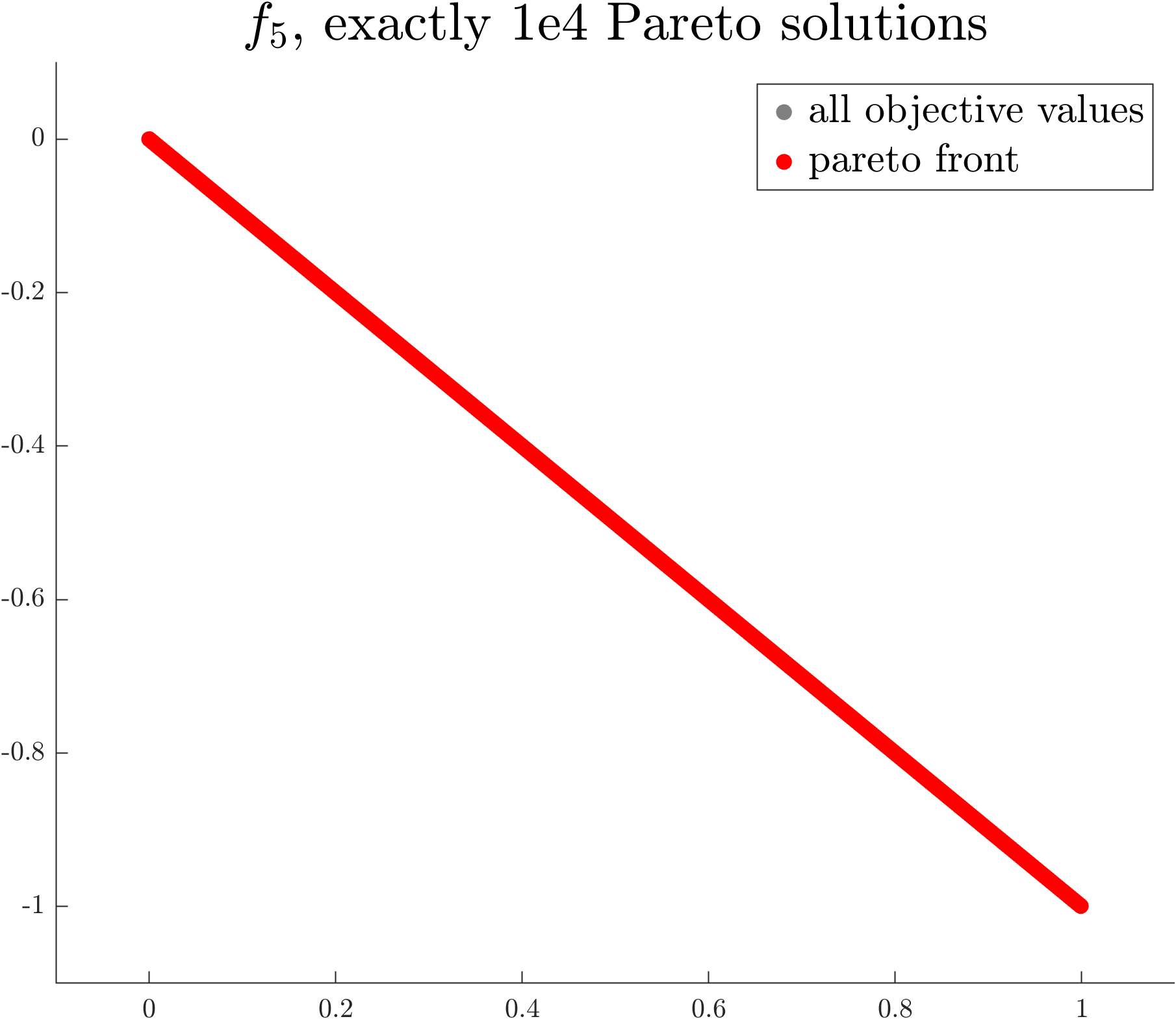}\\
   \end{tabular}\\
    \begin{tabular}{ccc}
      \includegraphics[width=0.2\linewidth]{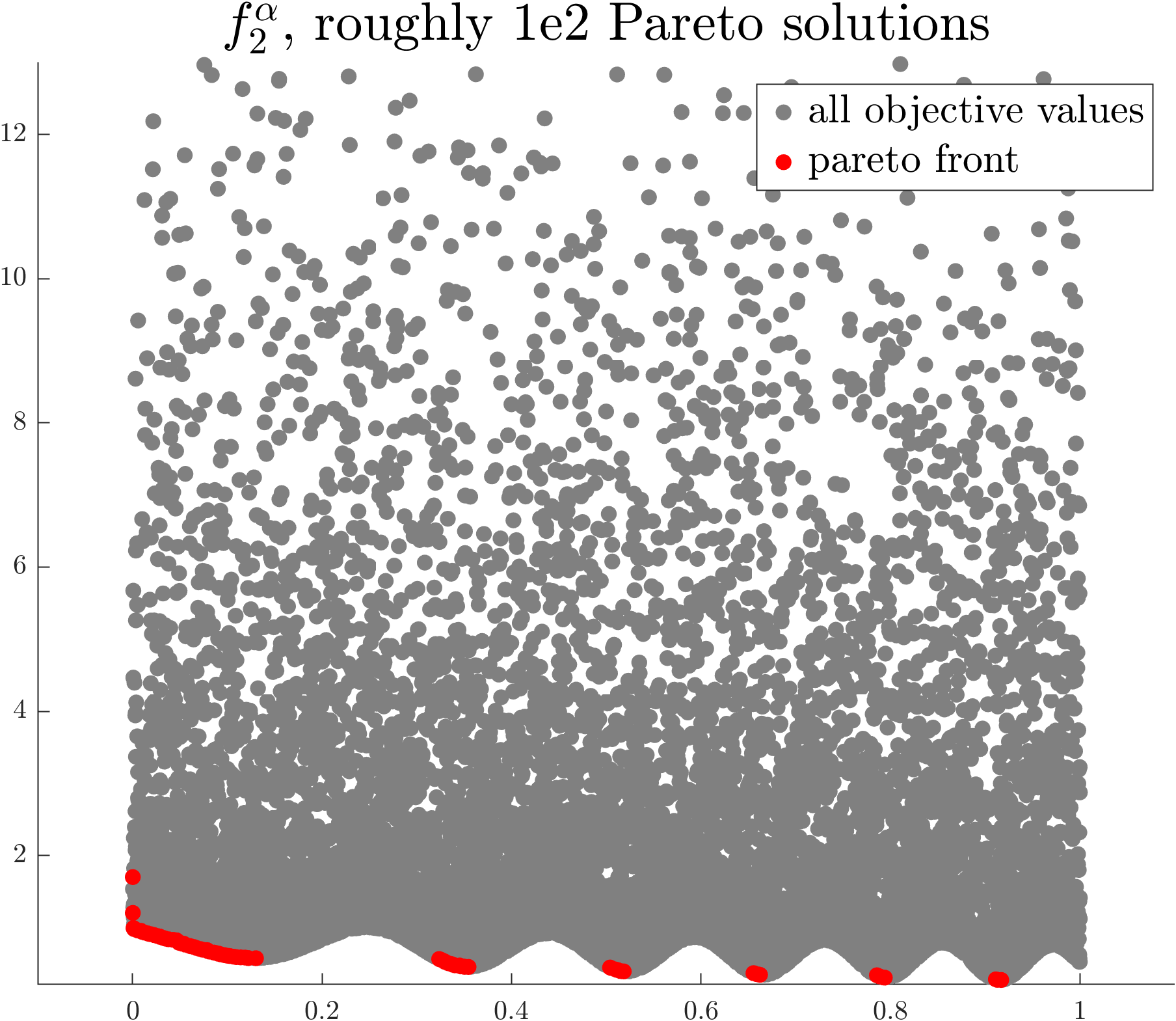}   & \includegraphics[width=0.2\linewidth]{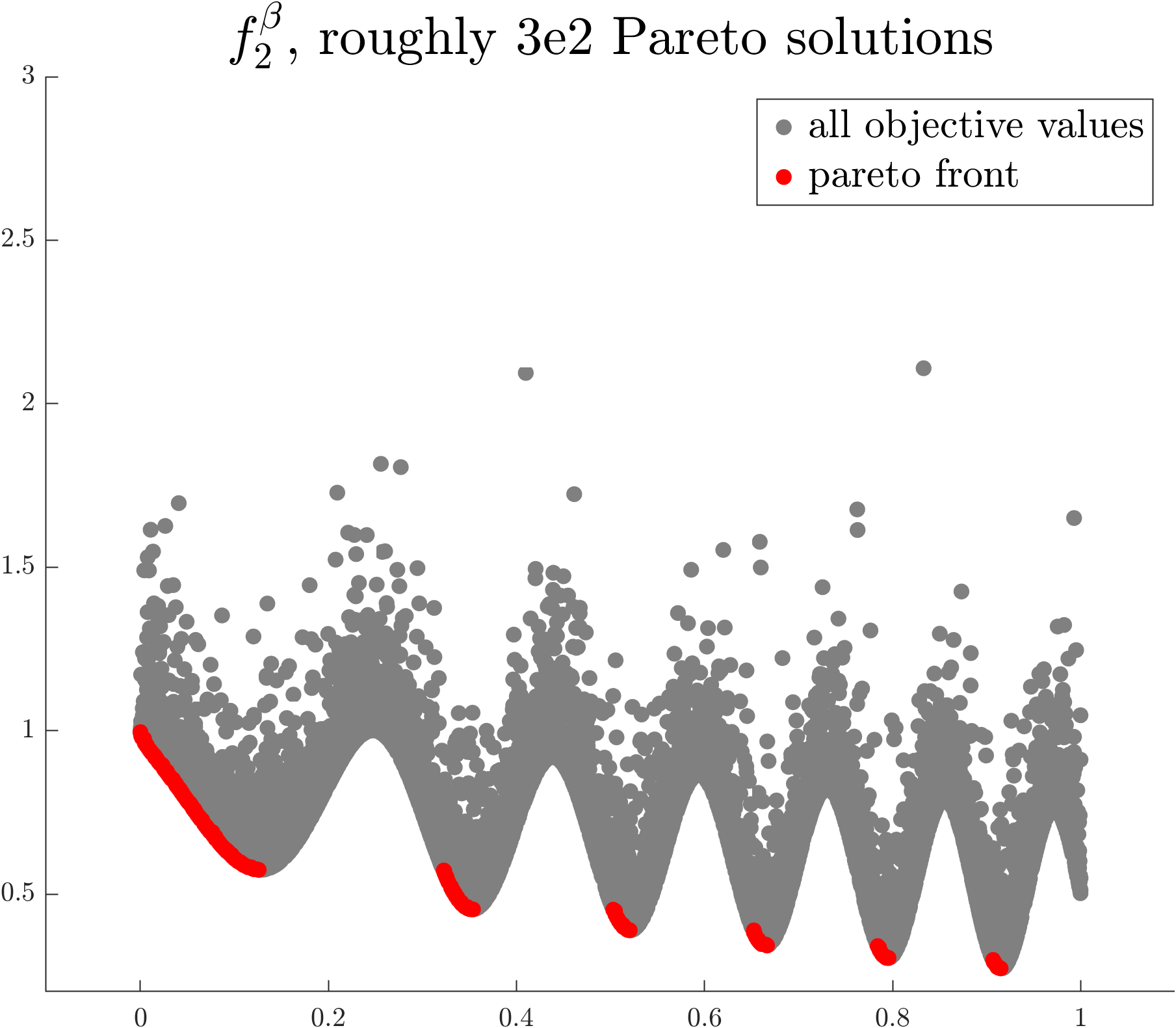}   & \includegraphics[width=0.2\linewidth]{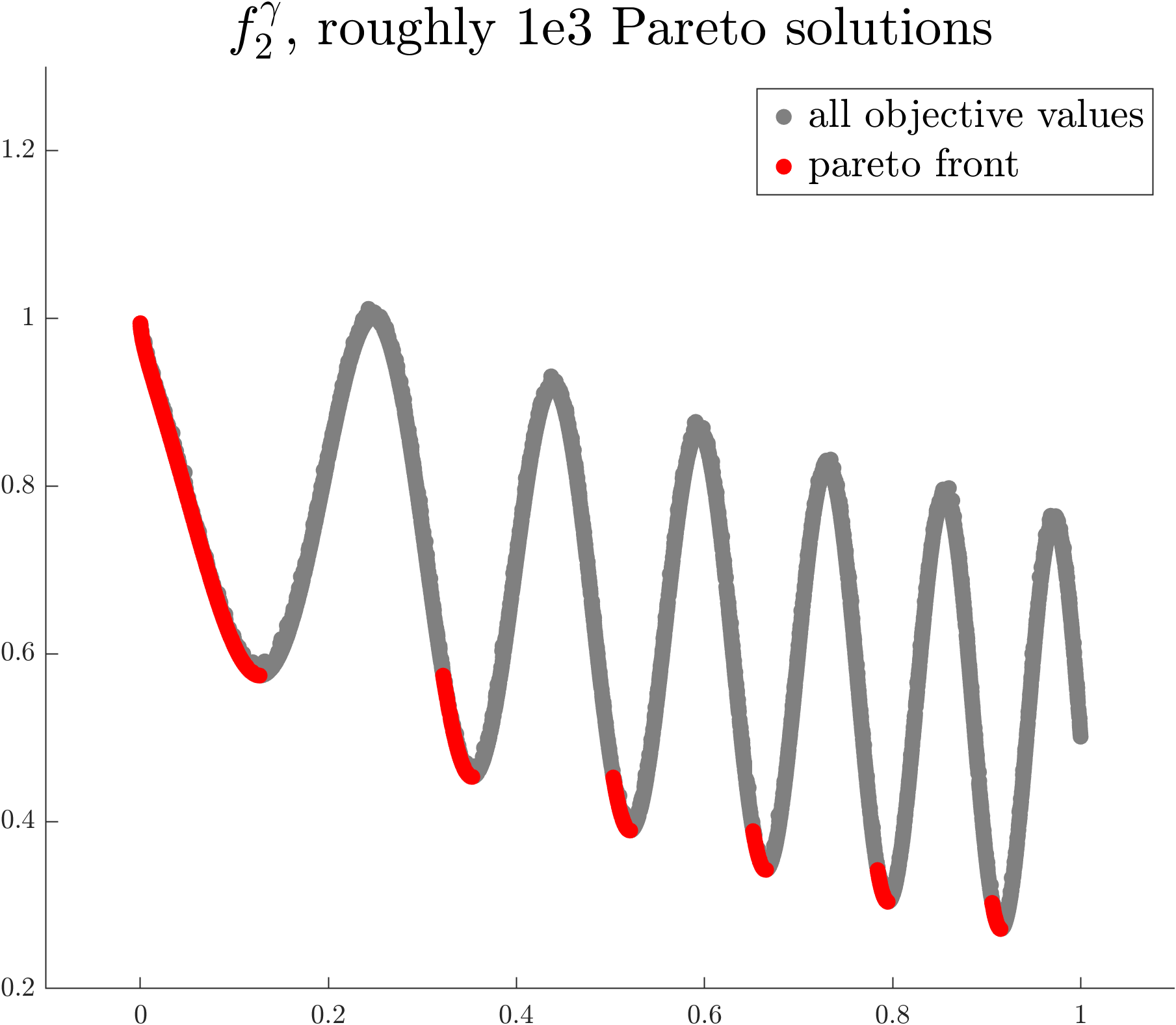}   \\
      \includegraphics[width=0.2\linewidth]{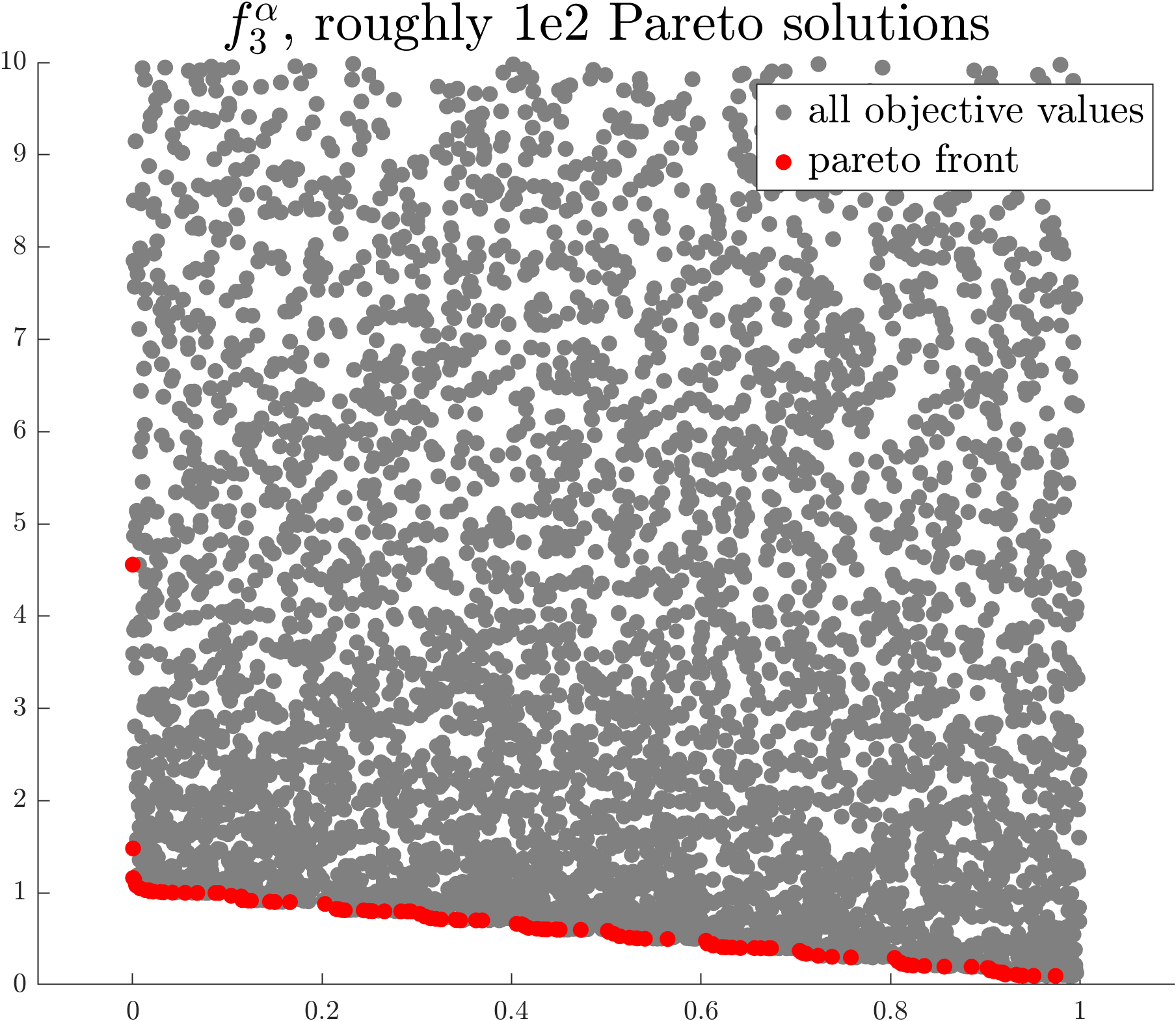}   & \includegraphics[width=0.2\linewidth]{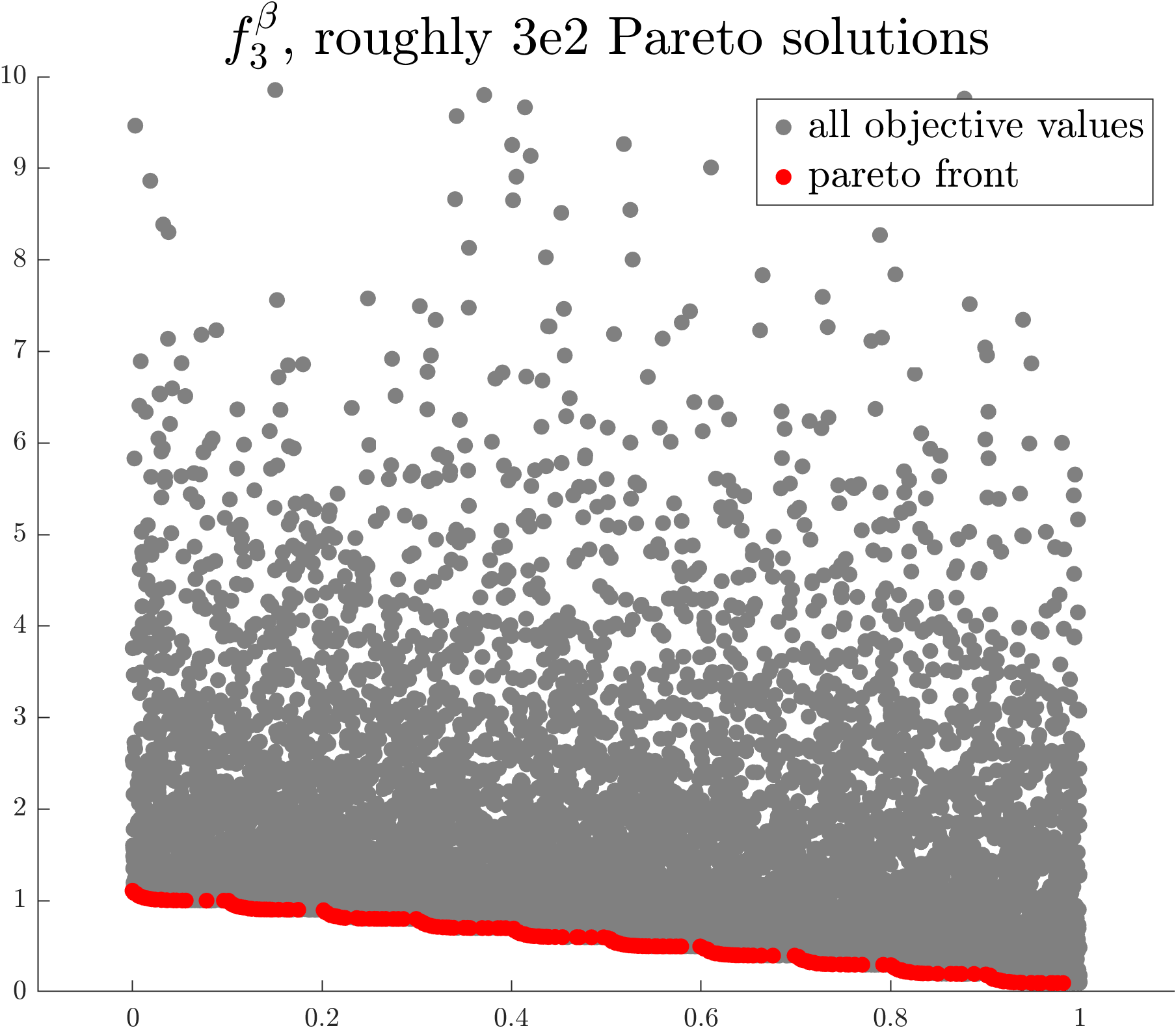}   & \includegraphics[width=0.2\linewidth]{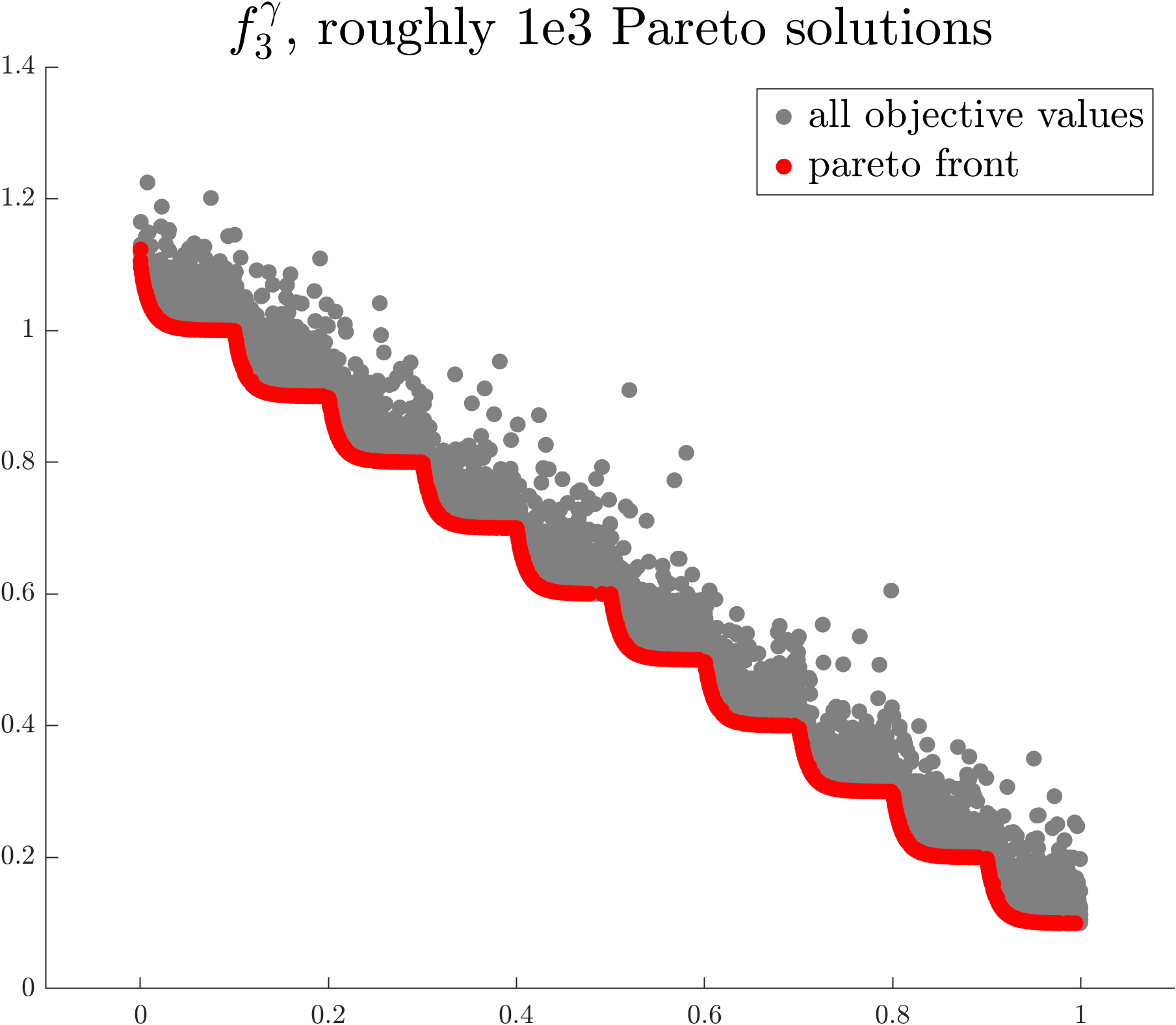}   \\
      \includegraphics[width=0.2\linewidth]{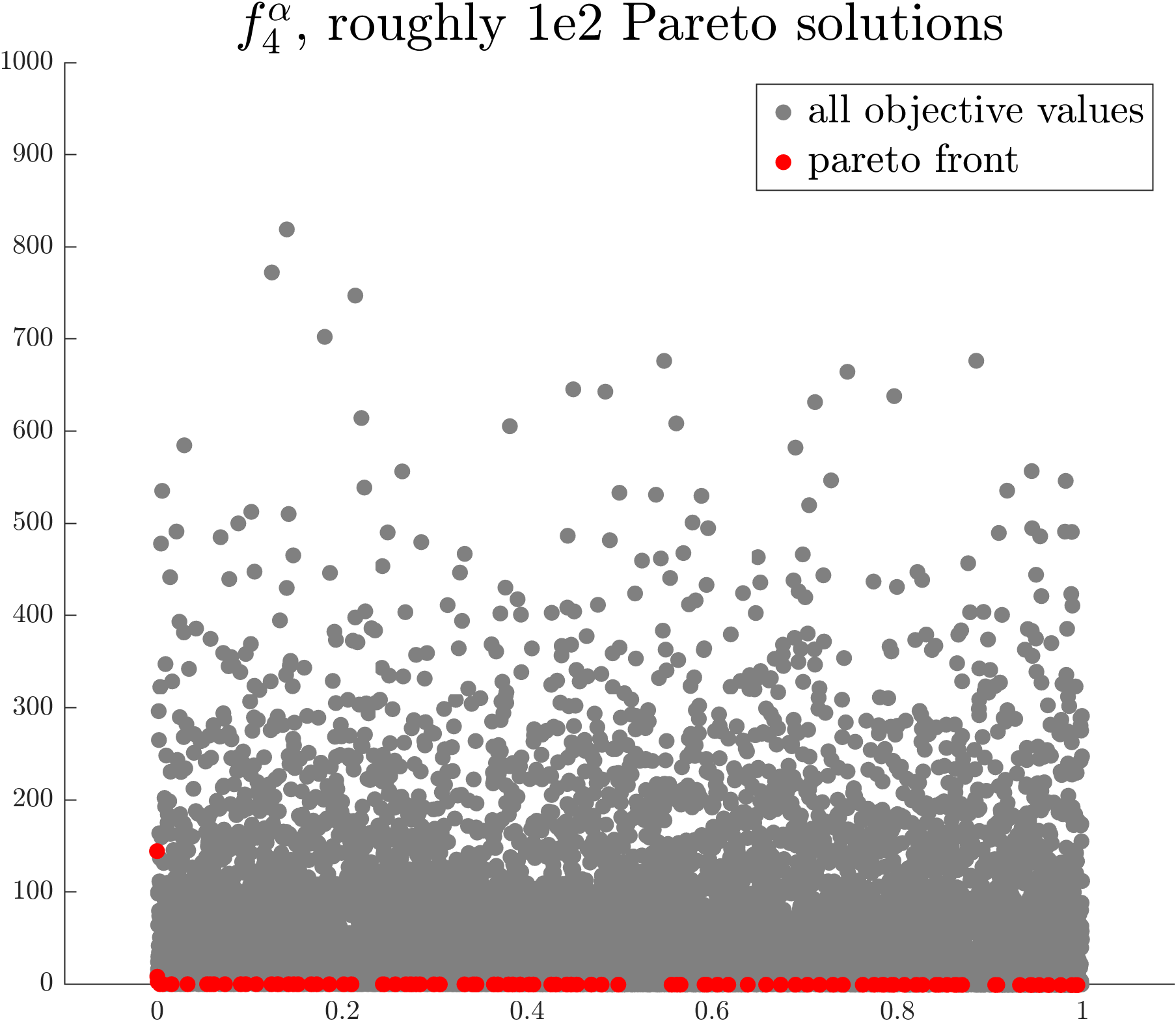}   & \includegraphics[width=0.2\linewidth]{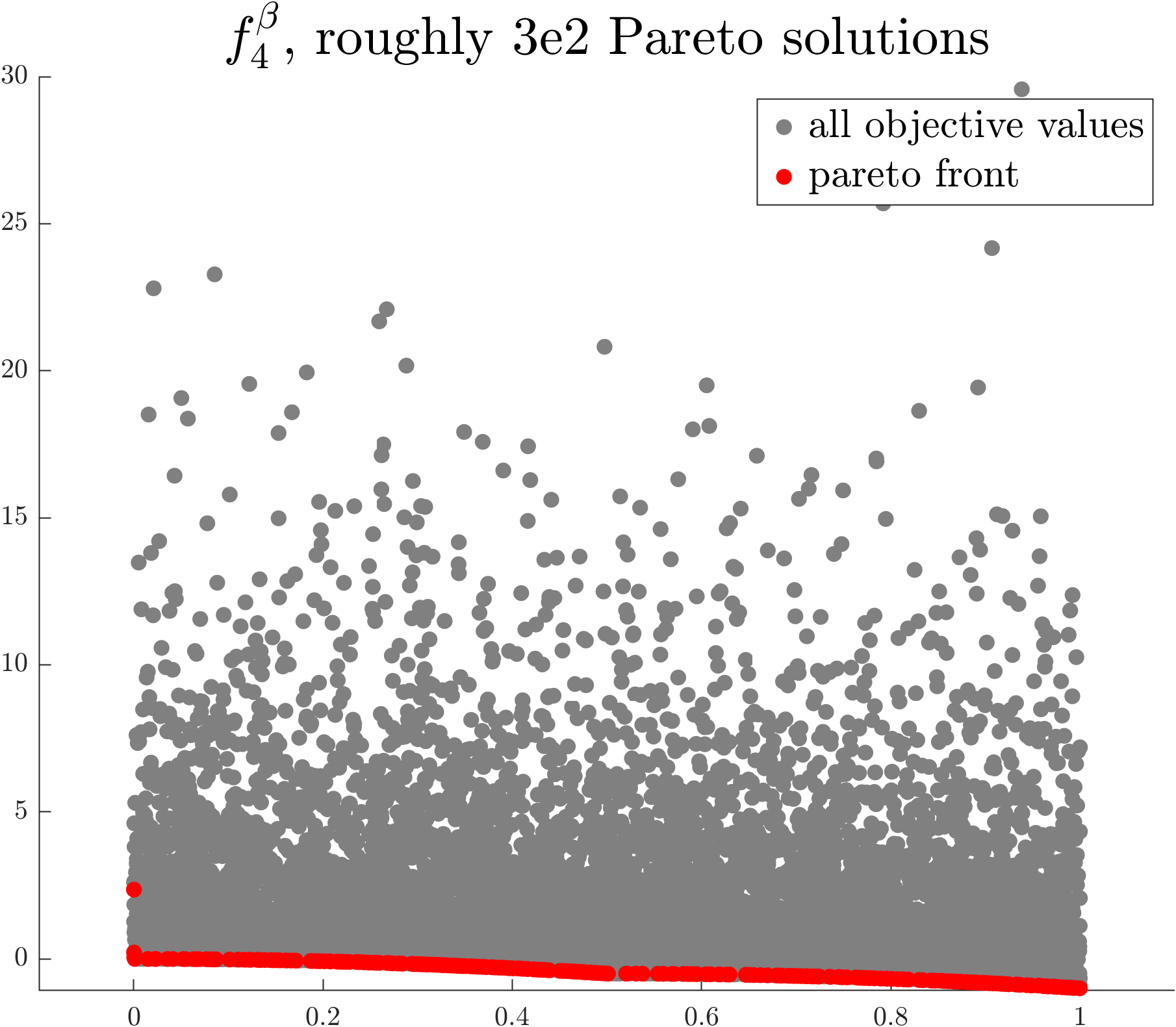}   & \includegraphics[width=0.2\linewidth]{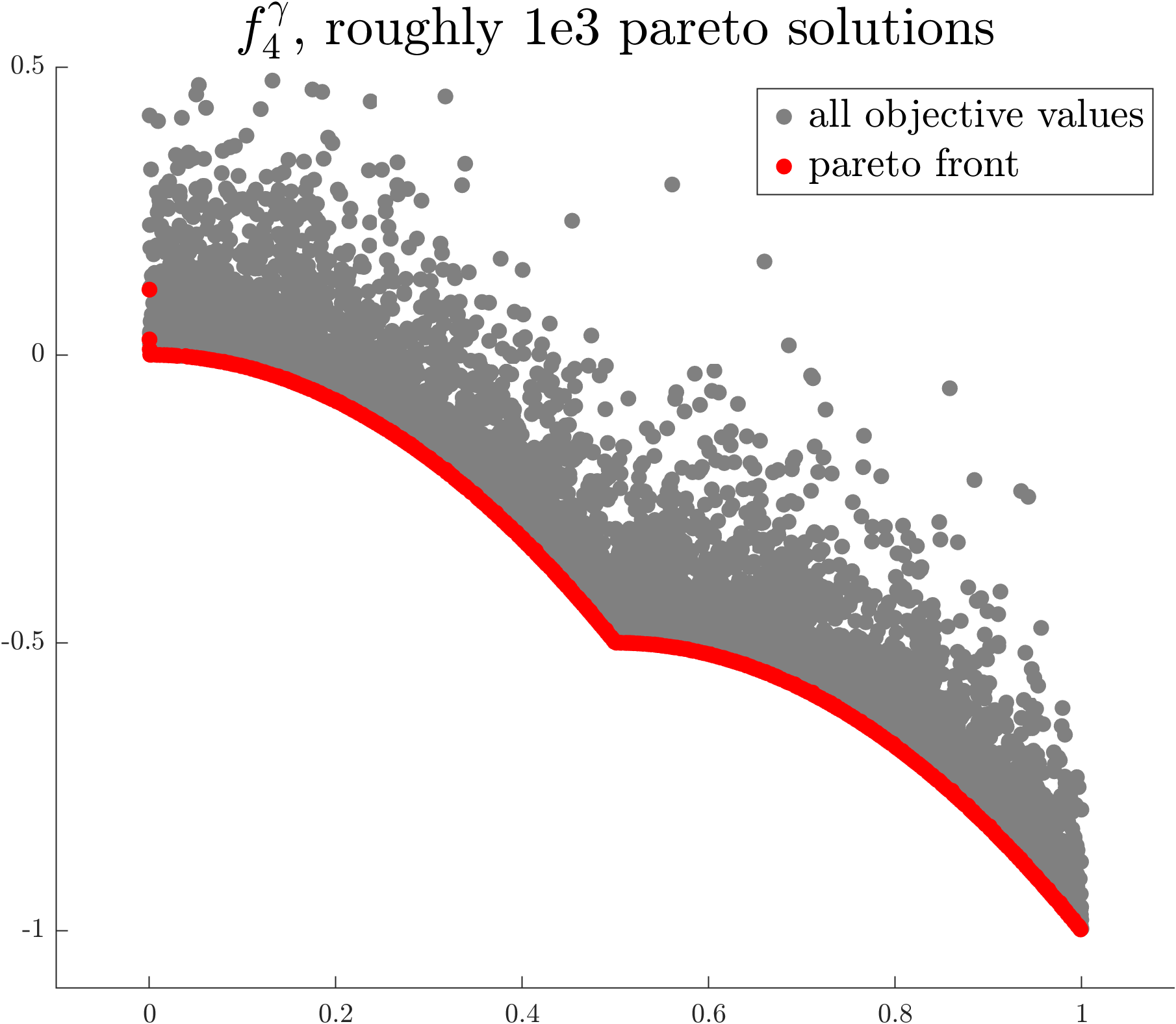}   \\
    \end{tabular}
    \caption{Proposed problems in 2d with the number of Pareto solutions in a sample of $10^4$ feasible solutions.}
    \label{fig:problems}
\end{figure*}

\begin{figure}
    \centering
    \includegraphics[width=.7\linewidth]{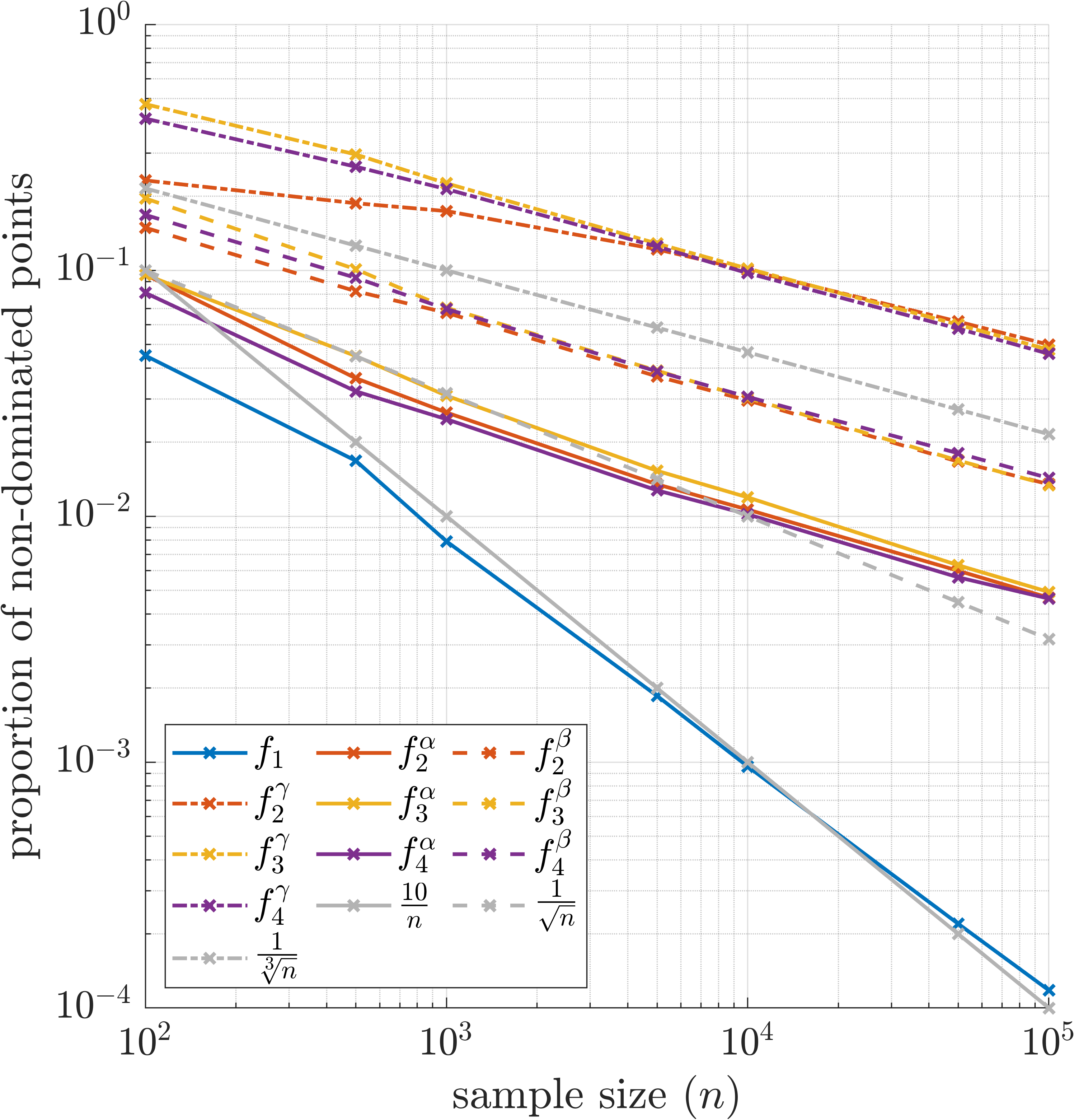}
    \caption{Relationship between sample size and number of non-dominated points for the proposed functions.}
    \label{fig:proportions}
\end{figure}

We propose 11 problems that we believe capture reasonably well the different properties of the Pareto front that might influence the behaviour of the algorithms. They all share the $f_0$ property of being ``random'', i.e. returning values that do not depend on the position of the decision variable $x$, with the decision space being the $[0,1]^d$ hypercube. They are all bi-objective, with the form $f = [g_1,g_2]$, see their definitions in Table~\ref{tab:problems} and visualisations of their Pareto fronts in Figure~\ref{fig:problems}:
\begin{description}
    \item[$f_1$] is the \textit{same} as using $f_0$ in both objectives. The (\textit{exact}) Pareto set for this problem is a single point [0,0]. In a sample of $10^4$ feasible solutions, roughly $10$ are non-dominated. The proportion of non-dominated points w.r.t. the sample size can be found in Figure~\ref{fig:proportions}. 
    \item[$f_2^{\{\alpha,\beta,\gamma \}}$] has a \textit{disconnected} Pareto set. The parametrisation for the $\alpha$, $\beta$, and $\gamma$ variant was chosen such that the proportion of Pareto set solutions in a $10^4$ sample was roughly 0.01, 0.03, and 0.1, respectively (the same applies for the parametrisations of $f_3$ and $f_4$).
    \item[$f_3^{\{\alpha,\beta,\gamma \}}$] has a Pareto set with several \textit{convex "stairs"} or "kinks".
    \item[$f_4^{\{\alpha,\beta,\gamma \}}$] has a Pareto set with \textit{two distinct concave regions}, connected with a sharp point.
    \item[$f_5$] is a problem where the two objectives are \textit{perfectly correlated} (negatively). For this problem, all sampled feasible points are non-dominated.
\end{description}

In $f_2$-$f_4$, the added randomness in $g_2$ has the form of (differently scaled) square of a standard normal distribution. The resulting sets of objective values, in our opinion, resemble more the MOO problems found in the literature than when using a uniform distribution (which would look like a ``band'' above the Pareto front). The scaling parameter $r$ lets us control the density of points near the Pareto front. The different shapes of the Pareto front, together with the different scaling parameters, produced a diverse range of correlation (conflict) structures between the two objectives (see values of $\rho$ in Table~\ref{tab:problems}).

Regarding selected parametrisations of functions $f_2$, $f_3$, $f_4$, we analyse the scaling and stability of the proportion of non-dominated points for every proposed function over multiple repetitions in Figure~\ref{fig:proportions}. Here, we can find that the chosen values of $r$ produce variants for $f_2$-$f_4$ that display similar dependence of proportion of non-dominated points on sample size (behaving roughly as the inverse of the cube root of the sample size). For $f_1$, behaves more like the inverse of the sample size (which is expected as $f_1$ has a single ``true'' Pareto front point).  

To summarise, the functions and their parametrisation were chosen with the aim of answering the following \textit{questions}:
\begin{itemize}
    \item Does the shape of the Pareto Front influence the type and/or severity of the SB?
    \item Does the proportion of non-dominated points w.r.t. the sample size influence the type and/or severity of the SB?
\end{itemize}

\subsection{Data collected from algorithms' runs}
We should stress that after running an algorithm on these problems, we are \textbf{not} interested in the ``quality'' of the non-dominated sets in terms of the objective functions' values: as the problems are random, all algorithms will have the ``same distribution'' of these values. What we are interested in is ``where the algorithm has searched'' within the decision space. Thus, to study this, we will focus on two datasets:
\begin{itemize}
    \item The set of decision variables $x$ that correspond to the \textit{non-dominated points over the whole run} of the algorithm ($X_P$).
    \item The set of decision variables $x$ that constitute the \textit{final population of the algorithm} ($X_L$).
\end{itemize}

\subsection{Proposed methodology of testing for SB in the MOO setting} An algorithm under investigation is run on the 11 functions with a predefined computation budget $B$ $n_r$ times. For run $i$, the $X_L^i$ and $X_P^i$ sets are saved per function. 

We are interested in whether the algorithm, when initialised uniformly over the search space\footnote{All algorithms implemented in PlatEMO enforce such initialisation, potentially overruling the choices originally made by algorithm designers. See also a disclaimer in Section~\ref{sect:discuss}.}, has ``preferred'' parts of the space where it searches more, or if it searches the space ``uniformly''. This question can be answered in multiple ways. 

\subsubsection{BIAS toolbox} One option for testing this behaviour is to use the BIAS~\cite{bib:BIAS} toolbox. We sample a single random point for each of the $\{X_P^1,\dots,X_P^{n_r}\}$ or $\{X_L^1,\dots,X_L^{n_r}\}$ sets independently, getting $n_r$ points in total. In the BIAS toolbox, these selected points are then subject to several (36 per dimension) tests for uniformity. This gives us a number of rejections that should correspond to ``how far from uniform'' the distribution of the selected points is. 

\subsubsection{Chi-squared test} The second option is to use a more simplistic (and computationally less demanding) methodology \cite{ibehej2025investigation} based on a chi-squared test. The selection of points is the same as when using the BIAS toolbox. The points are then aggregated into a $1d$ vector, divided into a given number (e.g., 20) of bins of equal size, and the size of these bins is compared to the expected size (from the uniform distribution). The result is a $p$-value of the chi-squared test.

\begin{figure}[!tb]
    \centering
    \begin{tikzpicture}[scale=2.5]
        \draw[->] (0,0) node[below]{0} -- (1,0) node[below]{1};
        \draw[->] (0,0) -- (0,1.15) node[pos=0.49, left=6pt, rotate=90]{Frequency};
        \draw[-] (0,0) -- (0,1.0) node[pos=1, left=1pt]{1};

        \fill[cyan!30, draw=black, line width=0.6pt] (0.05,0) rectangle (0.0,1.05);
        \fill[blue!30, draw=black, line width=0.6pt] (0.1,0) rectangle (0.05,0.97);
        \fill[blue!30, draw=black, line width=0.6pt] (0.15,0) rectangle (0.1,0.95);
        \fill[blue!30, draw=black, line width=0.6pt] (0.2,0) rectangle (0.15,1.1);
        \fill[blue!30, draw=black, line width=0.6pt] (0.25,0) rectangle (0.2,1.03);
        \fill[blue!30, draw=black, line width=0.6pt] (0.3,0) rectangle (0.25,0.9);
        \fill[blue!30, draw=black, line width=0.6pt] (0.35,0) rectangle (0.3,1.11);
        \fill[blue!30, draw=black, line width=0.6pt] (0.4,0) rectangle (0.35,0.89);
        \fill[blue!30, draw=black, line width=0.6pt] (0.45,0) rectangle (0.4,1);
        \fill[cyan!30, draw=black, line width=0.6pt] (0.5,0) rectangle (0.45,1.13);
        \fill[cyan!30, draw=black, line width=0.6pt] (0.55,0) rectangle (0.5,1.07);
        \fill[blue!30, draw=black, line width=0.6pt] (0.6,0) rectangle (0.55,1);
        \fill[blue!30, draw=black, line width=0.6pt] (0.65,0) rectangle (0.6,1.05);
        \fill[blue!30, draw=black, line width=0.6pt] (0.7,0) rectangle (0.65,1.05);
        \fill[blue!30, draw=black, line width=0.6pt] (0.75,0) rectangle (0.7,0.9);
        \fill[blue!30, draw=black, line width=0.6pt] (0.8,0) rectangle (0.75,0.89);
        \fill[blue!30, draw=black, line width=0.6pt] (0.85,0) rectangle (0.8,0.91);
        \fill[blue!30, draw=black, line width=0.6pt] (0.9,0) rectangle (0.85,1.03);
        \fill[blue!30, draw=black, line width=0.6pt] (0.95,0) rectangle (0.9,0.87);
        \fill[cyan!30, draw=black, line width=0.6pt] (1,0) rectangle (0.95,1.1);

        \draw[dashed, gray] (0,1) -- (1,1);

        \coordinate (binLB) at ($(0.0,1.05)!0.5!(0.05,1.05)$);
        \coordinate (binLC) at ($(0.45,1.13)!0.5!(0.5,1.13)$);
        \coordinate (binRC) at ($(0.5,1.07)!0.5!(0.55,1.07)$);
        \coordinate (binRB) at ($(0.95,1.1)!0.5!(1,1.1)$);

        \node[font=\scriptsize] at (binLB |- 0,1.25) {${}_B^L$};
        \node[font=\scriptsize] at ($(binLC |- 0,1.25) + (-0.01,0)$) {${}_C^L$};
        \node[font=\scriptsize] at ($(binRC |- 0,1.25) + (0.01,0)$) {${}_C^R$};
        \node[font=\scriptsize] at (binRB |- 0,1.25) {${}_B^R$};
    \end{tikzpicture}
    \hspace{1em}
    \begin{tikzpicture}[scale=2.5]
        \draw[->] (0,0) node[below]{0} -- (1,0) node[below]{1};
        \draw[->] (0,0) -- (0,1.7) node[pos=0.35, left=6pt, rotate=90]{Frequency};
        \draw[-] (0,0) -- (0,1.0) node[pos=1, left=1pt]{1};

        \fill[cyan!30, draw=black, line width=0.6pt] (0.05,0) rectangle (0.0,1.65);
        \fill[blue!30, draw=black, line width=0.6pt] (0.1,0) rectangle (0.05,0.9);
        \fill[blue!30, draw=black, line width=0.6pt] (0.15,0) rectangle (0.1,0.95);
        \fill[blue!30, draw=black, line width=0.6pt] (0.2,0) rectangle (0.15,1.0);
        \fill[blue!30, draw=black, line width=0.6pt] (0.25,0) rectangle (0.2,0.93);
        \fill[blue!30, draw=black, line width=0.6pt] (0.3,0) rectangle (0.25,0.9);
        \fill[blue!30, draw=black, line width=0.6pt] (0.35,0) rectangle (0.3,0.91);
        \fill[blue!30, draw=black, line width=0.6pt] (0.4,0) rectangle (0.35,0.79);
        \fill[blue!30, draw=black, line width=0.6pt] (0.45,0) rectangle (0.4,0.9);
        \fill[cyan!30, draw=black, line width=0.6pt] (0.5,0) rectangle (0.45,0.93);
        \fill[cyan!30, draw=black, line width=0.6pt] (0.55,0) rectangle (0.5,0.97);
        \fill[blue!30, draw=black, line width=0.6pt] (0.6,0) rectangle (0.55,0.9);
        \fill[blue!30, draw=black, line width=0.6pt] (0.65,0) rectangle (0.6,0.95);
        \fill[blue!30, draw=black, line width=0.6pt] (0.7,0) rectangle (0.65,0.95);
        \fill[blue!30, draw=black, line width=0.6pt] (0.75,0) rectangle (0.7,0.9);
        \fill[blue!30, draw=black, line width=0.6pt] (0.8,0) rectangle (0.75,0.89);
        \fill[blue!30, draw=black, line width=0.6pt] (0.85,0) rectangle (0.8,0.91);
        \fill[blue!30, draw=black, line width=0.6pt] (0.9,0) rectangle (0.85,1.03);
        \fill[blue!30, draw=black, line width=0.6pt] (0.95,0) rectangle (0.9,0.97);
        \fill[cyan!30, draw=black, line width=0.6pt] (1,0) rectangle (0.95,1.67);

        \draw[dashed, gray] (0,1) -- (1,1);

        \coordinate (binLB2) at ($(0.0,1.65)!0.5!(0.05,1.65)$);
        \coordinate (binLC2) at ($(0.45,0.93)!0.5!(0.5,0.93)$);
        \coordinate (binRC2) at ($(0.5,0.97)!0.5!(0.55,0.97)$);
        \coordinate (binRB2) at ($(0.95,1.67)!0.5!(1,1.67)$);

        \node[font=\scriptsize] at (binLB2 |- 0,1.75) {${}_B^L$};
        \node[font=\scriptsize] at ($(binLC2 |- 0,1.10) + (-0.01,0)$) {${}_C^L$};
        \node[font=\scriptsize] at ($(binRC2 |- 0,1.10) + (0.01,0)$) {${}_C^R$};
        \node[font=\scriptsize] at (binRB2 |- 0,1.75) {${}_B^R$};
    \end{tikzpicture}
    \caption{Examples of typical histograms of the aggregation of the results, with four bins of interest indicated in a different colour, see Section~\ref{sect:visual} for explanation. 
    }
    \label{fig:histogram}
\end{figure}
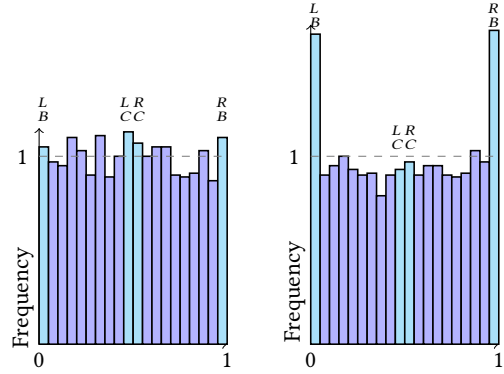

\begin{figure}[!tb]
    \centering
    \includegraphics[width=0.48\textwidth]{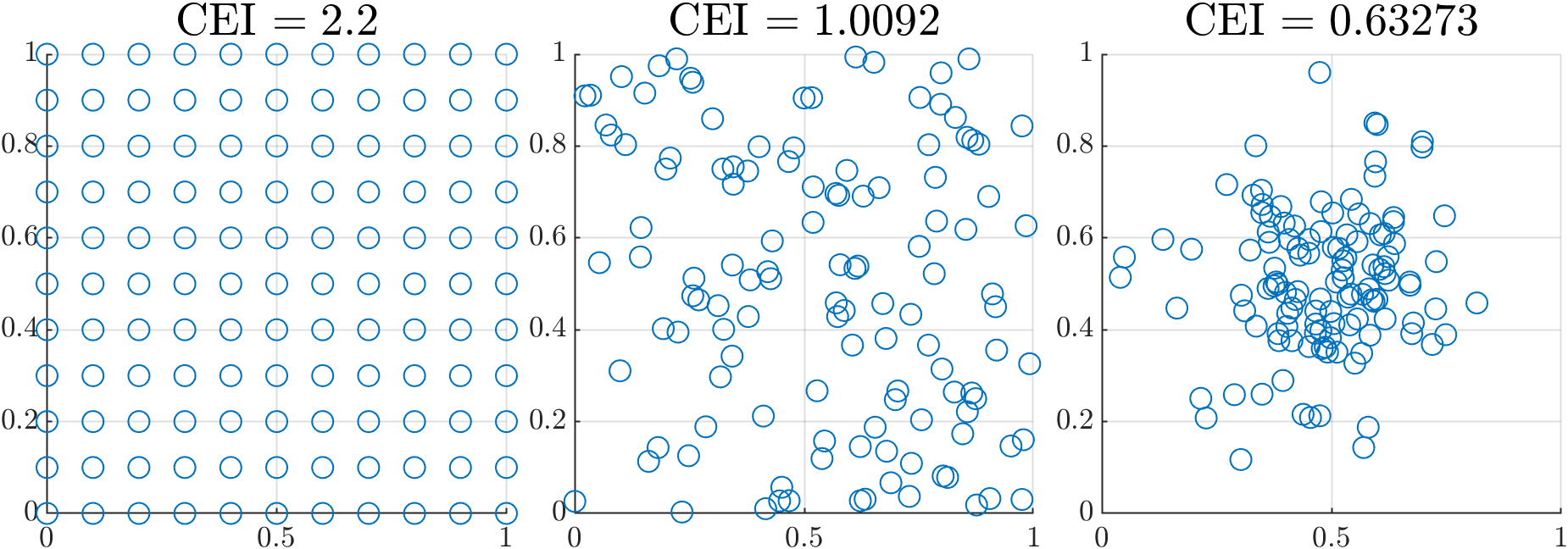}
    \caption{Visual representation of CEI values, see Section~\ref{sect:cluster} for explanation.}
    \label{fig:CEI}
\end{figure}

\subsubsection{Binsize inspection}\label{sect:visual}
Using all points in $\{X_P^1,\dots,X_P^{n_r}\}$ or \\ $\{X_L^1,\dots,X_L^{n_r}\}$, we aggregate them into a single sample
\[
\mathcal{X} = \bigcup_{i=1}^{n_r} X^{i},
\]
and construct a one-dimensional histogram by partitioning the interval $[0,1]$ into $K$ equally sized bins
\[
B_k = \left[\frac{k-1}{K}, \frac{k}{K}\right), \quad k=1,\dots,K .
\]

Let $n_k$ denote the number of points falling into bin $B_k$, and let $N = |\mathcal{X}|$ be the total number of aggregated points.  
The relative bin size is defined as
\[
\text{binsize}_k = \frac{n_k}{N/K},
\]
where $N/K$ is the expected count under a uniform distribution.

We focus on boundary bins
\[
\text{binsize}_B^L = \text{binsize}_1, \qquad
\text{binsize}_B^R = \text{binsize}_K,
\]
and, for even $K$, the two central bins
\[
\text{binsize}_C^L = \text{binsize}_{K/2}, \qquad
\text{binsize}_C^R = \text{binsize}_{K/2+1}.
\]

Values of $\text{binsize}_B^{L,R} \gg 1$ indicate bound bias, while elevated $\text{binsize}_C^{L,R}$ indicate centre bias, corresponding to the two most common structural bias patterns identified in~\cite{bib:BIAS}.  
Typical histograms are shown in Figure~\ref{fig:histogram}: the left illustrates an approximately uniform distribution, while the right exhibits bound bias with $\text{binsize}_B^L = 1.65$ and $\text{binsize}_B^R = 1.67$.

\subsubsection{Clustering}\label{sect:cluster} The last quantity we'll be interested in is the amount of clustering/concentration in the individual runs (i.e., if the search within an individual run converges to a specific location in the search space). For this, we will use the $X_L$ data from the individual runs and compute the Clark-Evans Index (CEI)~\cite{clark1954distance}. Values $\text{CEI} >> 1$ signify a grid-like distribution of points, $\text{CEI} \approx 1$ signal random (uniform) distribution of points, while $\text{CEI} < 1$ means some sort of clustering. A visual representation of different CEI values is shown on Figure~\ref{fig:CEI}. 

\begin{figure}[!tb]
    \centering
    \includegraphics[width=\linewidth]{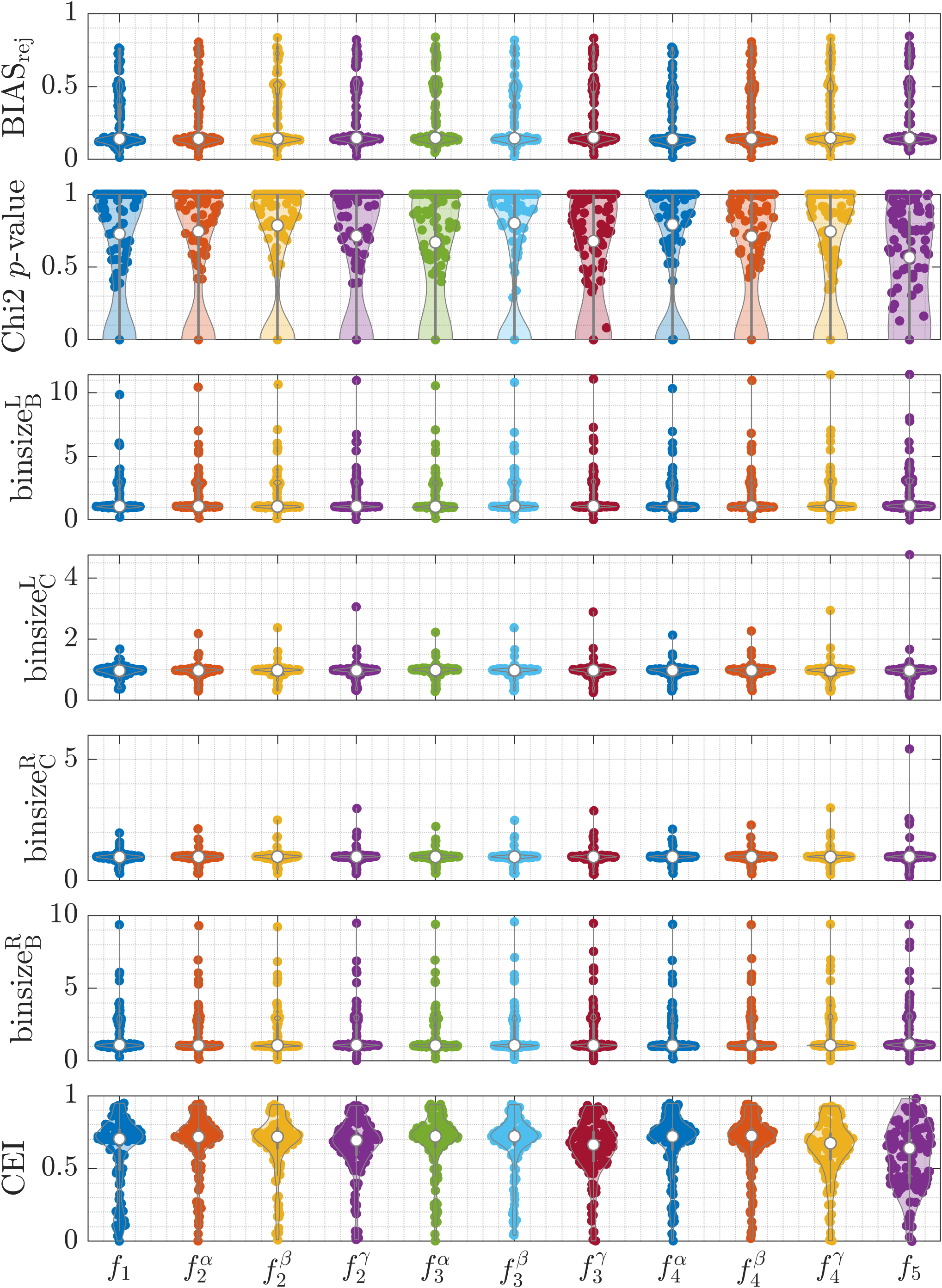}
    \caption{Distribution of values of the tracked parameters used for detection of structural bias, $d=10$ (The corresponding figure for $d=2$ can be found in the supplementary material).} 
    \label{fig:violin_all}
\end{figure}

\begin{figure*}[!tb]
    \centering
    \includegraphics[width=0.8\linewidth,trim=1mm 1mm 4mm 1mm,clip]{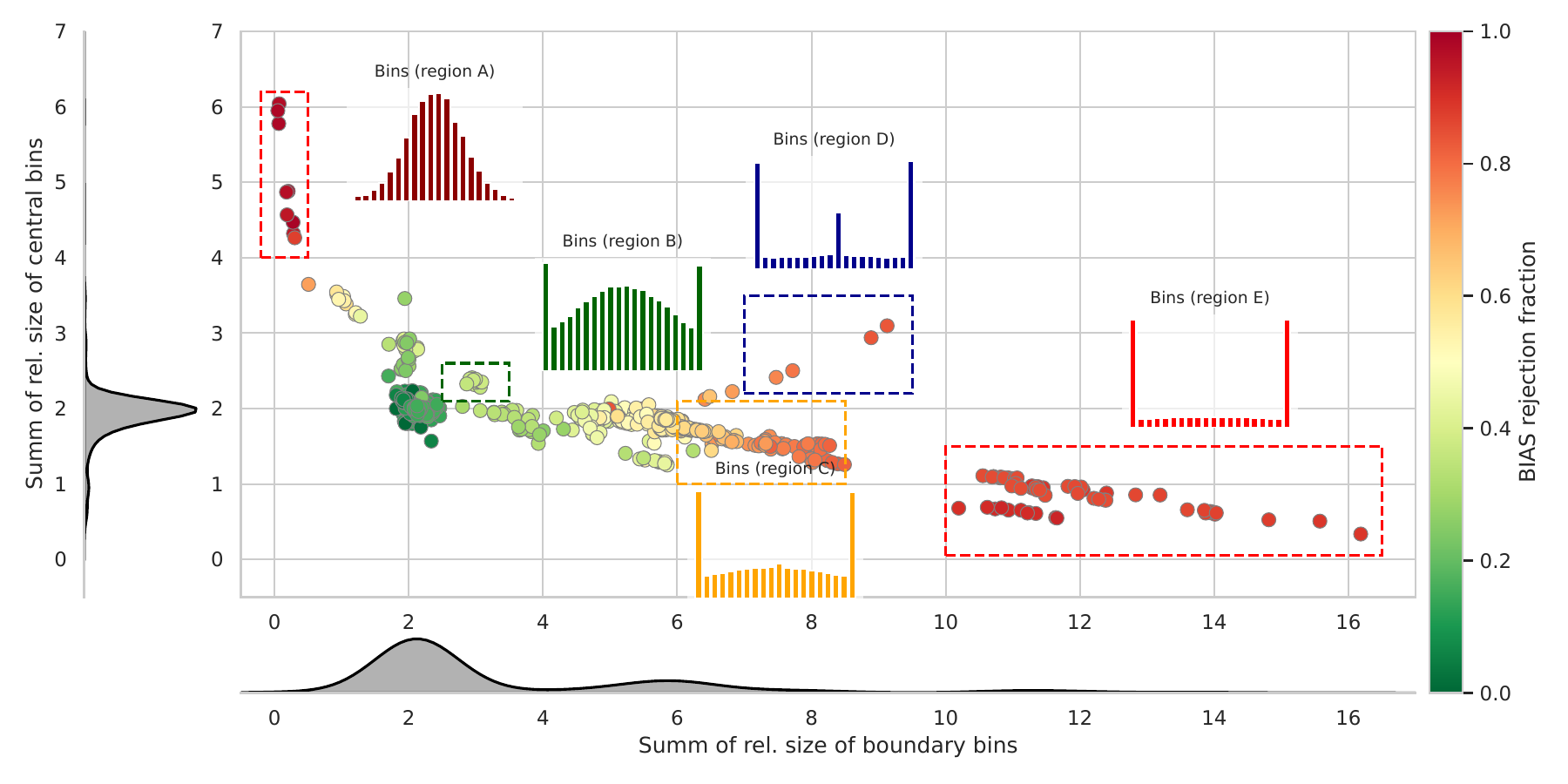}
    \caption{The sum of relative sizes of boundary versus centre bins with different regions showing different distributions and fraction of test rejections (colour). The dark green region around ($2,2$) is made up of algorithms that show less structural bias.} 
    \label{fig:regions}
\end{figure*}

\subsection{Experimental setup}
To test the above-described methodology, we used the PlatEMO platform \cite{tian2017platemo}, which contains more than 300 implementations of both standard and state-of-the-art (single and multi-objective) algorithms. From this large collection, we selected \textit{120 algorithms} that fit the simple criteria: being fit for continuous multi-objective bound-constrained problems, without the use of surrogates, so that the experiments could be done in a reasonable time, The full list of algorithms can be found in the Supplementary material.

We used 10 repetitions for the estimation of SB using the BIAS toolbox and the chi-squared test. In the results, we report the average number of rejections from the BIAS toolbox (over the 10 repetitions) as $\text{BIAS}_\text{rej}$, and the geometric average of the $p$-values from the chi-squared test, scaled by a factor of $e$ \cite{vovk2020combining}.

The experiments on the 11 functions were performed with the default (PlatEMO) settings for methods, except for the population size (set to 100) and number of iterations (set to 300), in dimensions $D=\{2,10\}$, and with $n_r = 100$ repetitions. The budget for the computation was thus $30{,}000$ function evaluations.

In PlatEMO, the default setting for algorithm initialisation is random uniform over the search space. During the execution of the algorithm, the default setting for enforcing bound constraints is saturation \cite{Kononova2024_tiobr}.

The results of the experiments (also with histograms and population distribution plots for all algorithm-function-dimension triplets), together with the code used to run them, are available at a Zenodo repository\footnote{\url{https://doi.org/10.5281/zenodo.18380058}}.

\section{Discussion}\label{sect:discuss}
\textbf{Disclaimer}: The conclusions regarding the presence, type and strength of structural bias in this study are limited to the specific algorithm implementations provided in the PlatEMO package and to their default parameter settings, as specified therein. We therefore encourage the authors of the individual algorithms considered in this study to conduct a more comprehensive analysis of their methods, explicitly accounting for the effects of algorithmic parameterisations. \textit{Detailed results} per algorithm can be found in the Supplementary material and in our Zenodo repository.

\subsection{Results aggregated across PlatEMO} 
Figure~\ref{fig:violin_all} shows the violin plots for the tracked parameters of all 120 algorithms, aggregated over the 11 functions. Here, several observations can be made. On aggregate, practically all tracked parameters have (more or less pronounced) dependence on the proportion of non-dominated points - i.e., the extremes of binsizes increase, while the CEI decreases (signifying more clustering) when going from $f_1$ to $f_5$, or from the $\alpha$ to $\gamma$ variants of $f_2$-$f_4$. 

Figure~\ref{fig:regions} summarises the five ``typical'' bias types that were found for multiple algorithms and which can be inferred from the binsizes: in region A are the ``pure'' centre-biased methods; in region E the ``pure'' bound-biased methods; and in regions B, C, and D the mix of the two with slightly different characteristics. Note that many of the algorithms lie in the region around $(2,2)$, which is the point where the relative boundary and centre bin sizes are equal to the expected bin size of a uniform distribution (and as such, unbiased). The spread in the boundary bin sizes ($x$-axis) is larger than that in the centre bin sizes ($y$-axis). We hypothesise that the high prevalence of taller bins at the domain bounds observed across PlatEMO arises from the enforced saturation of infeasible solutions at those bounds. This observation underscores the \textit{importance of bound constraint handling} in the MOO setting, rather than being an issue limited to SOO~\cite{Kononova2024_tiobr}. 

\begin{figure}
    \centering
    \begin{subfigure}[b]{0.245\linewidth}
        \centering
        \includegraphics[width=1\linewidth]{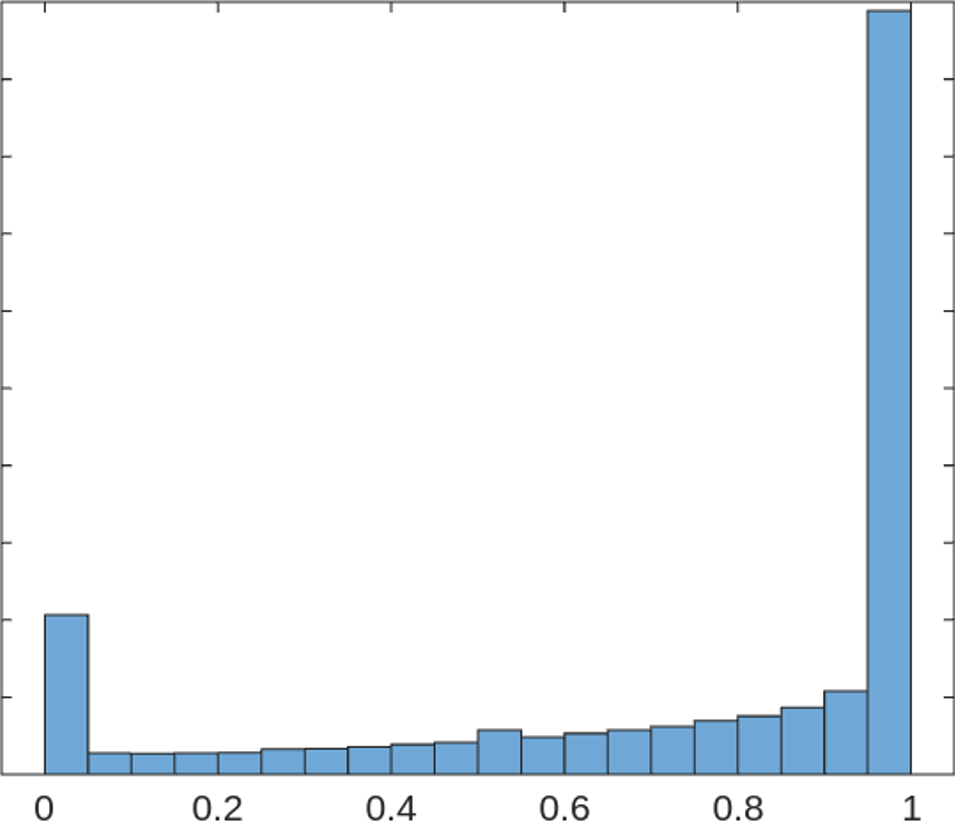}
        \caption{1-heavy}
    \end{subfigure}\hfill
    \begin{subfigure}[b]{0.245\linewidth}
        \centering
        \includegraphics[width=1\linewidth]{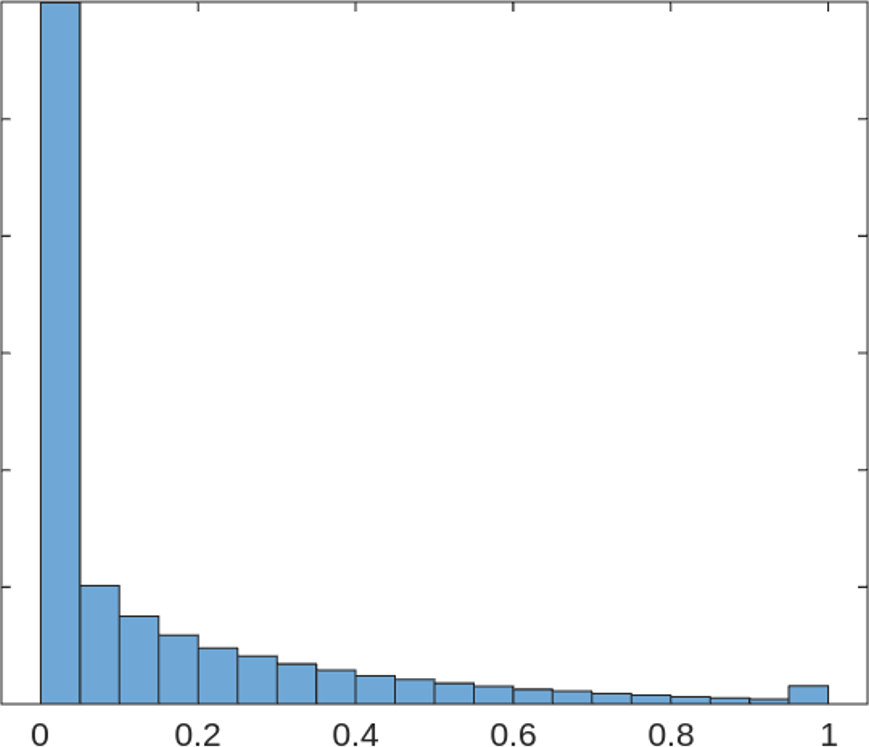}
        \caption{0-heavy}
    \end{subfigure}\hfill
    \begin{subfigure}[b]{0.245\linewidth}
        \centering
        \includegraphics[width=1\linewidth]{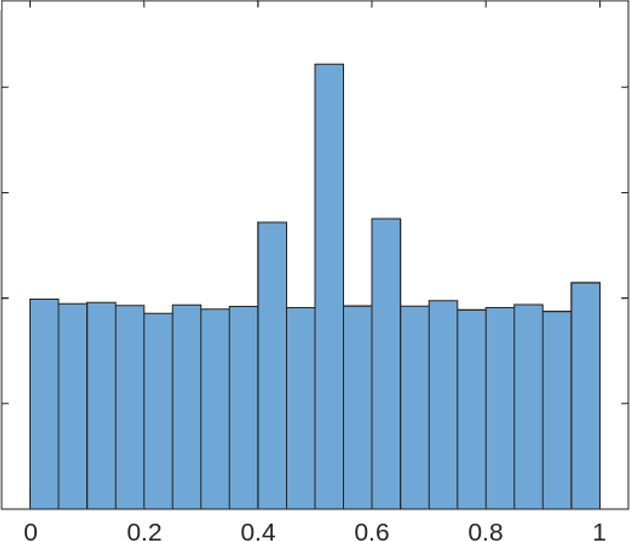}
        \caption{trident}
    \end{subfigure}\hfill
    \begin{subfigure}[b]{0.245\linewidth}
        \centering
        \includegraphics[width=1\linewidth]{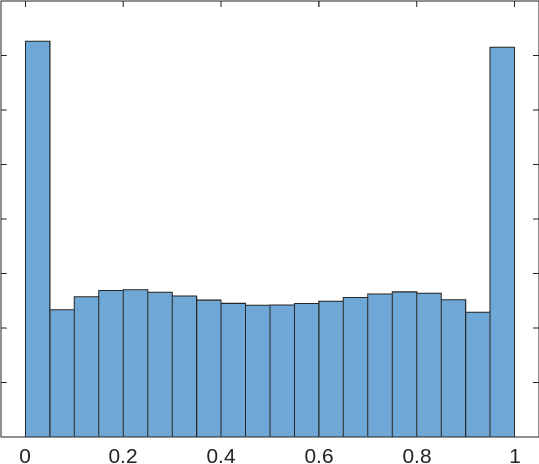}
        \caption{wave+bounds}
    \end{subfigure}\hfill
    \caption{Examples of unexpected histogram shapes.}
    \label{fig:unex}
\end{figure}

\begin{figure*}[!tb]
    \centering
    \includegraphics[width=0.31\textwidth,trim=0mm 0mm 0mm 0mm,clip]{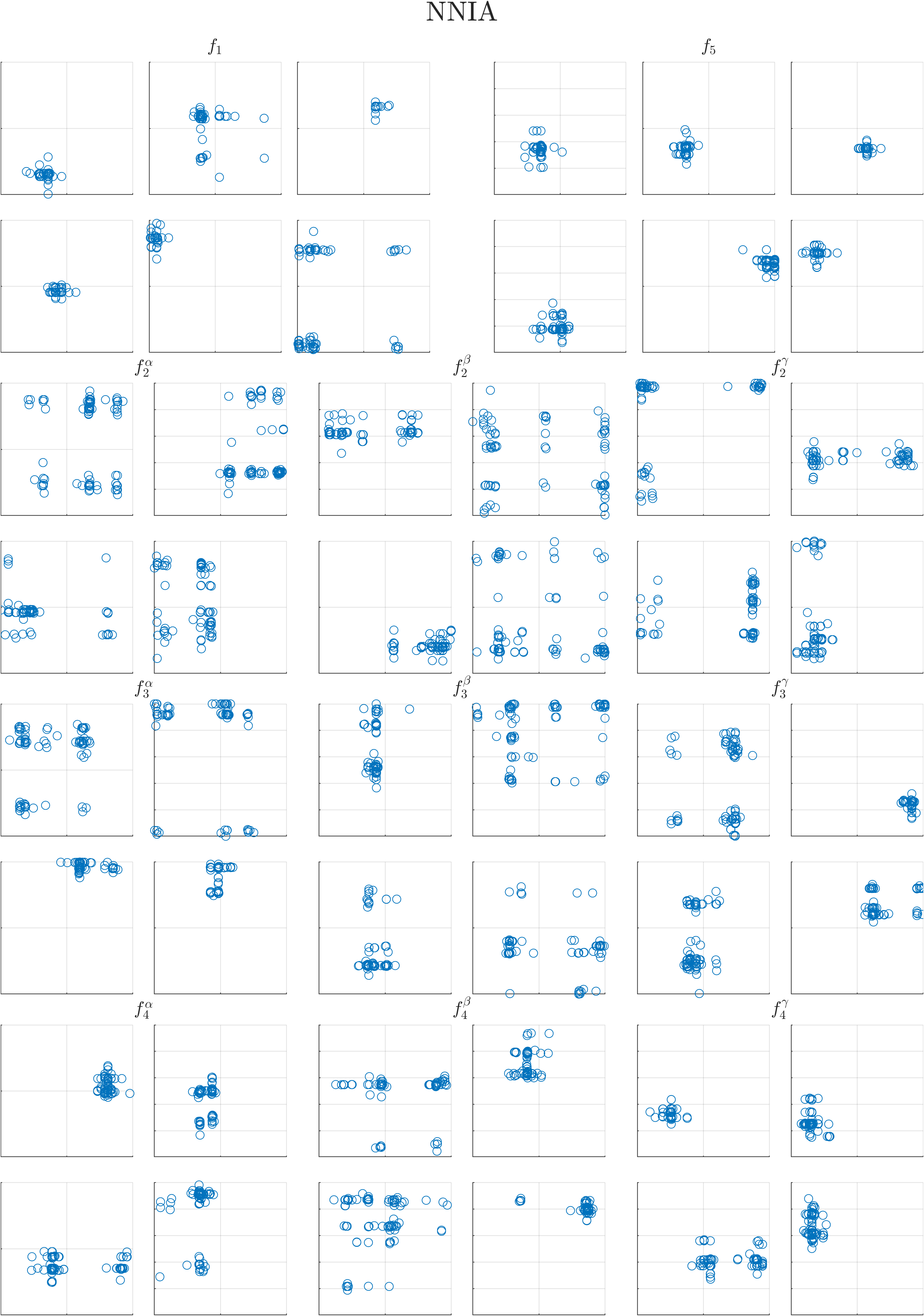}\label{fig:nnia}
    \includegraphics[width=0.31\textwidth,trim=0mm 0mm 0mm 0mm,clip]{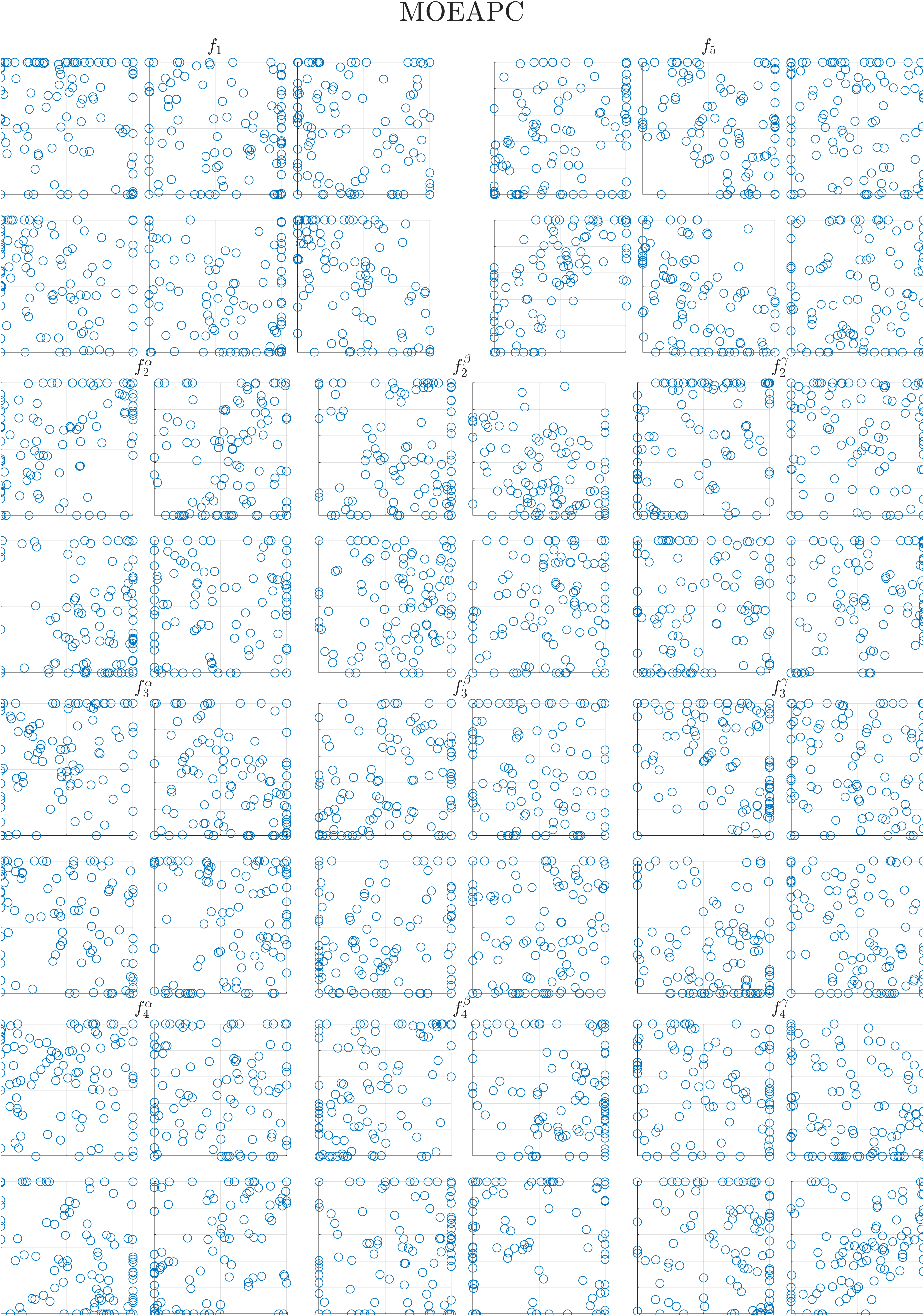}\label{fig:moeapc}
    \includegraphics[width=0.31\textwidth,trim=0mm 0mm 0mm 0mm,clip]{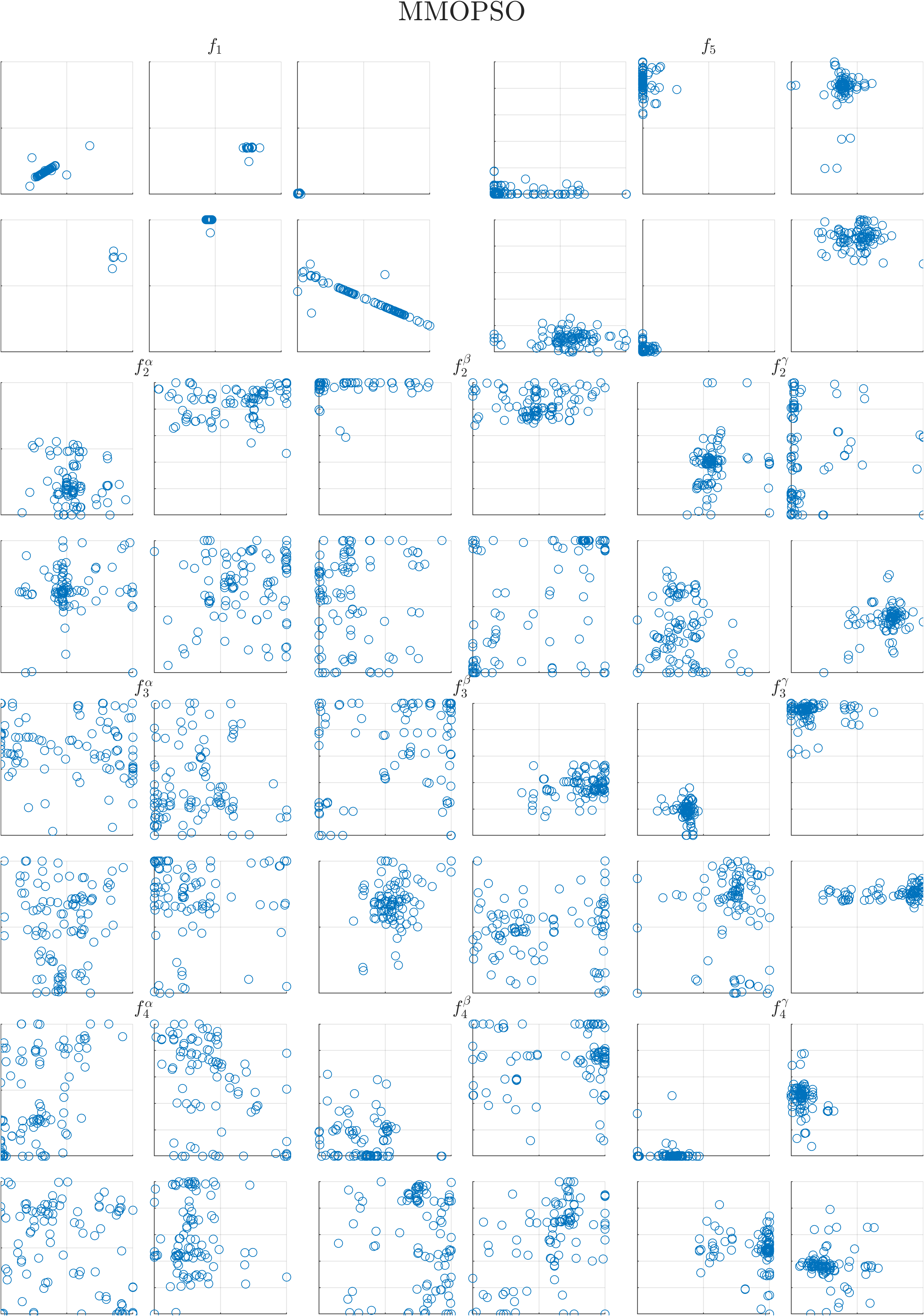}\label{fig:mmpso}
    \caption{Locations of points in the final population of the PlatEMO implementations of algorithms NNIA, MOEAPC, MMOPSO (the first two dimensions of the $d=10$ experiment). Each small figure within corresponds to one run of the algorithm on functions, organised in rows and columns: 2x3 runs on $f_1$ (rows 1-2, left) and 2x3 runs on $f_5$ (rows 1-2, right), followed by 2x2 runs on parameterised functions $f_2^\alpha$ (rows 3-4, left ), $f_2^\beta$ (rows 3-4, middle ), $f_2^\gamma$ (rows 3-4, right), $f_3^\alpha$ (rows 5-6, left ), $f_3^\beta$ (rows 5-6, middle), $f_3^\gamma$ (rows 5-6, right), $f_4^\alpha$ (rows 7-8, left ), $f_4^\beta$ (rows 7-8, middle ), $f_4^\gamma$ (rows 7-8, right).}\label{fig:final_pops}
\end{figure*}

\subsection{Interesting cases of individual algorithms} 
\begin{table*}[!ht]
\caption{Various cases of dependence of CEI on function. \vspace{-3mm}}\label{tab:cei} 
\resizebox{0.85\linewidth}{!}{
\begin{tabular}{lrccccccccccc} \hline
Algorithm      & $d$ & $f_1$                                           & $f_2^\alpha$                                    & $f_2^\beta$                                     & $f_2^\gamma$                                    & $f_3^\alpha$                                    & $f_3^\beta$                                     & $f_3^\gamma$                                    & $f_4^\alpha$                                    & $f_4^\beta$                                     & $f_4^\gamma$                                    & $f_5$                                           \\ \hline
MOEADM2M & 2                     & \cellcolor[HTML]{4CB86F}7.31E$-$01              & \cellcolor[HTML]{3E3A7A}\color{white}3.15E$-$01 & \cellcolor[HTML]{3F3275}\color{white}2.88E$-$01 & \cellcolor[HTML]{402A70}\color{white}2.59E$-$01 & \cellcolor[HTML]{33658B}\color{white}4.52E$-$01 & \cellcolor[HTML]{3C4A85}\color{white}3.64E$-$01 & \cellcolor[HTML]{402A70}\color{white}2.42E$-$01 & \cellcolor[HTML]{39A77B}6.81E$-$01              & \cellcolor[HTML]{39A77B}6.72E$-$01              & \cellcolor[HTML]{39A77B}6.70E$-$01              & \cellcolor[HTML]{43095A}\color{white}1.41E$-$01 \\
MSEA     & 2                     & \cellcolor[HTML]{7ECF56}8.36E$-$01              & \cellcolor[HTML]{3C4A85}\color{white}3.77E$-$01 & \cellcolor[HTML]{3C4A85}\color{white}3.63E$-$01 & \cellcolor[HTML]{2B788C}\color{white}5.18E$-$01 & \cellcolor[HTML]{3C4A85}\color{white}3.64E$-$01 & \cellcolor[HTML]{3E3A7A}\color{white}3.22E$-$01 & \cellcolor[HTML]{365F8B}\color{white}4.44E$-$01 & \cellcolor[HTML]{3C4A85}\color{white}3.69E$-$01 & \cellcolor[HTML]{3D4280}\color{white}3.39E$-$01 & \cellcolor[HTML]{2E728C}\color{white}4.93E$-$01 & \cellcolor[HTML]{46B373}7.21E$-$01              \\
NSGAIII  & 2                     & \cellcolor[HTML]{58C366}7.78E$-$01              & \cellcolor[HTML]{58C366}7.87E$-$01              & \cellcolor[HTML]{52BE6A}7.55E$-$01              & \cellcolor[HTML]{4CB86F}7.41E$-$01              & \cellcolor[HTML]{58C366}7.82E$-$01              & \cellcolor[HTML]{58C366}7.79E$-$01              & \cellcolor[HTML]{4CB86F}7.36E$-$01              & \cellcolor[HTML]{58C366}7.83E$-$01              & \cellcolor[HTML]{52BE6A}7.70E$-$01              & \cellcolor[HTML]{52BE6A}7.59E$-$01              & \cellcolor[HTML]{58C366}7.74E$-$01              \\
MMOPSO   & 10                    & \cellcolor[HTML]{440154}\color{white}3.35E$-$02 & \cellcolor[HTML]{279788}6.20E$-$01              & \cellcolor[HTML]{279788}6.26E$-$01              & \cellcolor[HTML]{2B788C}\color{white}5.20E$-$01 & \cellcolor[HTML]{33A27F}6.54E$-$01              & \cellcolor[HTML]{33A27F}6.58E$-$01              & \cellcolor[HTML]{3C4A85}\color{white}3.65E$-$01 & \cellcolor[HTML]{2D9C84}6.33E$-$01              & \cellcolor[HTML]{33A27F}6.56E$-$01              & \cellcolor[HTML]{3D4280}\color{white}3.44E$-$01 & \cellcolor[HTML]{3D4280}\color{white}3.26E$-$01 \\
CTSEA    & 10                    & \cellcolor[HTML]{2E728C}\color{white}4.99E$-$01 & \cellcolor[HTML]{52BE6A}7.57E$-$01              & \cellcolor[HTML]{52BE6A}7.51E$-$01              & \cellcolor[HTML]{40AD77}7.06E$-$01              & \cellcolor[HTML]{52BE6A}7.68E$-$01              & \cellcolor[HTML]{52BE6A}7.51E$-$01              & \cellcolor[HTML]{2D9C84}6.38E$-$01              & \cellcolor[HTML]{52BE6A}7.63E$-$01              & \cellcolor[HTML]{4CB86F}7.46E$-$01              & \cellcolor[HTML]{2D9C84}6.38E$-$01              & \cellcolor[HTML]{365F8B}\color{white}4.43E$-$01 \\
NSGAII   & 10                    & \cellcolor[HTML]{40AD77}6.90E$-$01              & \cellcolor[HTML]{46B373}7.15E$-$01              & \cellcolor[HTML]{39A77B}6.87E$-$01              & \cellcolor[HTML]{21918C}\color{white}6.00E$-$01 & \cellcolor[HTML]{46B373}7.17E$-$01              & \cellcolor[HTML]{40AD77}6.91E$-$01              & \cellcolor[HTML]{2B788C}\color{white}5.28E$-$01 & \cellcolor[HTML]{46B373}7.18E$-$01              & \cellcolor[HTML]{39A77B}6.90E$-$01              & \cellcolor[HTML]{2B788C}\color{white}5.12E$-$01 & \cellcolor[HTML]{3D4280}\color{white}3.54E$-$01 \\ 
MOEADDU &	10	&\cellcolor[HTML]{58C366}7.89E$-$01&
\cellcolor[HTML]{8ED250}8.57E$-$01	&\cellcolor[HTML]{8ED250}8.66E$-$01	&\cellcolor[HTML]{8ED250}8.66E$-$01	&\cellcolor[HTML]{5EC962}8.10E$-$01	&\cellcolor[HTML]{5EC962}8.01E$-$01	&\cellcolor[HTML]{5EC962}8.05E$-$01	&\cellcolor[HTML]{5EC962}8.05E$-$01	&\cellcolor[HTML]{6ECC5C}8.24E$-$01	&\cellcolor[HTML]{AED844}9.10E$-$01	&\cellcolor[HTML]{EDE42B}9.80E$-$01\\
\hline
\end{tabular}
}
\end{table*}
There were also many cases of unexpected behaviour for different algorithms: see nonsymmetric or otherwise strange histograms in Figure~\ref{fig:unex}. 

Moreover, no strong relationship was observed between the different types of identified bias and the CEI. In Figure~\ref{fig:final_pops}, we present three algorithms exhibiting diverse behaviours; additional examples can be found in the Supplementary Material and the associated Zenodo repository.
\begin{itemize}
    \item NNIA was found to be close to unbiased ($\text{BIAS}_\text{rej}\approx0.1$, Chi2 $p$-value $> 0.3 $, binsizes $\approx 1$), while having a very low value of CEI ($\approx 0.2$). 
    \item MOEAPC was found to be bound-biased ($\text{BIAS}_\text{rej}\approx0.5$, Chi2 $p$-value $\ll 0.01 $, binsize$_B^{L,R} \approx 3$) but with a high value of CEI ($\approx 0.9$). 
    \item MMOPSO was also found to be bound-biased ($\text{BIAS}_\text{rej}\approx0.4$, Chi2 $p$-value $\ll 0.01 $, binsize$_B^{L,R} \approx 2.5$), but with low values of CEI (and structural patterns such as points distributed along a straight line), which highly depended on the function. 
\end{itemize}

Among the tracked parameters, the CEI values displayed the largest dependence on the shapes of the objective functions. Several examples are shown in Table~\ref{tab:cei} - we see CEI values are very low for $f_2$, $f_3$ and $f_5$, but not for $f_1$ and $f_4$ (MOEADM2M); much lower for $f_1$ and for $f_5$, but falling for $f_2$-$f_4$ as the proportion of Pareto solutions increases, i.e. going from $\alpha$ to $\gamma$ (MMOPSO), etc.

Overall, the ``least biased'' algorithm with high values of CEI was MOEADDU, which, according to the authors, was designed with the intent of balancing convergence and diversity. 

\subsection{Further reflections on structural bias}
MOO adds a natural extra dimension and therefore perspective to SB and algorithm behaviour in general. Therefore, it becomes more difficult in the MOO to decouple fitness landscape influences from structural behaviour than it is in the SOO.

Overall, the proposed set of test problems and SB analysis tools consistently reveal non-uniform search behaviour across a large set of MOO algorithms. The different detection signals (BIAS tests, binsizes, CEI) broadly agree on the presence of structural bias.
However, the proposed methodology can still be improved by further investigating the link between bias measures and concrete algorithmic mechanisms (e.g., constraint handling like saturation) and by formalising robustness across frameworks beyond PlatEMO. The shape of the Pareto front does not meaningfully affect bias severity or type, but it affects clustering (CEI), while smoothness and convexity mainly modulate convergence patterns rather than introducing new SB patterns. The size (density) of the PF clearly matters: higher proportions of non-dominated points amplify boundary effects and clustering, making SB more visible. The $\rho$ correlation structure influences how strongly algorithms concentrate but not whether SB exists. Not all functions are strictly necessary for detecting bias presence, but the diverse set is useful to separate PF-density effects from geometric effects; more functions would mainly help stress-test generality rather than change conclusions. Different results across functions primarily mean that some observed behaviours are algorithm–PF-interaction effects (especially for clustering), while the underlying structural bias itself is largely algorithm-driven and consistent over all proposed functions.

While designing functions presented in this paper, we became increasingly more aware of a caveat for testing specialised algorithms via random functions, both in the SOO and MOO: we implicitly assume that a specialised algorithm makes reasonable steps on a function it is not aimed for, which is not the case (e.g. adaptive or learning-heavy algorithms). This requires further investigation. 

\section{Conclusions and Outlook}\label{sect:conclusions}
This paper introduced the first systematic extension of structural bias analysis to the multi-objective optimisation setting. By decoupling algorithmic behaviour from fitness-driven guidance through deliberately uninformative multi-objective test problems, we demonstrated that structural bias is not limited to single-objective optimisation but is also prevalent in multi-objective algorithms. The results indicate that many state-of-the-art MOO methods exhibit clear location-dependent search preferences in the search space, arising purely from algorithmic design choices such as selection, diversity preservation and constraint handling.

\paragraph{Contributions}
We proposed a suite of synthetic bi-objective problems with analytically controlled Pareto front properties and showed how existing SB detection methodologies can be adapted to multi-objective settings. Across a large set of algorithms implemented in PlatEMO, consistent deviations from uniform search behaviour were observed. Different Pareto front shapes do not seem to influence the severity or presence of structural bias. They do influence the amount of clustering (CEI) observed in individual runs.
We also identified groups of algorithms showing clear boundary bias, centre bias or a mix of both, with more variation and severity in the boundary bias variants. 

\paragraph{Future work}
Extending the methodology beyond the bi-objective case to higher-dimensional objective spaces is a natural next step, where dominance relations and diversity mechanisms may introduce different forms of structural bias. Further research is also needed to connect structural bias to performance outcomes on real benchmark problems, clarifying when bias is harmful, neutral or even potentially beneficial. From a methodological perspective, the SB detection techniques can still be refined and verified on additional MOO algorithm packages. Ultimately, incorporating behavioural analysis such as structural bias into standard benchmarking practice offers a more complete understanding of how multi-objective algorithms search and why they succeed or fail.


\putbib
\end{bibunit}

\clearpage
\onecolumn
\appendix

\section{Supplementary material}

\begin{bibunit}[plainnat]
\renewcommand{\bibname}{Appendix References} 
\nobibliography{biblio_appendix}

\nocite{*}

\subsection{Algorithms in the PlatEMO package considered in the paper}\label{sect:algorithms}
{\scriptsize
\begin{longtable}[!tb]{p{0.9cm}p{2.1cm}p{7cm}p{5.8cm}c} 
    \textbf{Reference} & \textbf{PlatEMO abbreviation} & \textbf{PlatEMO full name of the algorithm} & \textbf{Publication venue} & \textbf{Year}\\ \midrule
    \endfirsthead
    \textbf{Reference} & \textbf{PlatEMO abbreviation} & \textbf{PlatEMO full name of the algorithm} & \textbf{Publication venue} & \textbf{Year}\\ \midrule
    \endhead
    \cite{li2020weights} & \hyperref[AdaW]{AdaW} & Evolutionary algorithm with adaptive weights & Evolutionary Computation & 2020\\
    \cite{panichella2019adaptive} & \hyperref[AGEMOEA]{AGEMOEA} & Adaptive geometry estimation-based many-objective evolutionary algorithm & Proceedings of the Genetic and Evolutionary Computation Conference & 2019\\
    \cite{panichella2022improved} & \hyperref[AGEMOEAII]{AGEMOEAII} & Adaptive geometry estimation-based many-objective evolutionary algorithm II & Proceedings of the Genetic and Evolutionary Computation Conference & 2022\\
    \cite{jain2013evolutionary} & \hyperref[ANSGAIII]{ANSGAIII} & Adaptive NSGA-III & IEEE Transactions on Evolutionary Computation & 2013\\
    \cite{tian2017indicator} & \hyperref[ARMOEA]{ARMOEA} & Adaptive reference points based multi-objective evolutionary algorithm & IEEE Transactions on Evolutionary Computation & 2017\\
    \cite{li2015pareto} & \hyperref[BCEIBEA]{BCEIBEA} & Bi-criterion evolution based IBEA & IEEE Transactions on Evolutionary Computation & 2015\\
    \cite{li2015pareto} & \hyperref[BCEMOEAD]{BCEMOEAD} & Bi-criterion evolution based MOEA/D & IEEE Transactions on Evolutionary Computation & 2015\\
    \cite{liu2021handling} & \hyperref[BiCo]{BiCo} & Bidirectional coevolution constrained multiobjective evolutionary algorithm & IEEE Transactions on Cybernetics & 2021\\
    \cite{sun2022multistage} & \hyperref[C3M]{C3M} & Constraint, multiobjective, multi-stage, multi-constraint evolutionary algorithm & IEEE Transactions on Evolutionary Computation & 2022\\
    \cite{zou2021dual} & \hyperref[CAEAD]{CAEAD} & Dual-population evolutionary algorithm based on alternative evolution and degeneration & Information Sciences & 2021\\
    \cite{hua2018clustering} & \hyperref[CAMOEA]{CAMOEA} & Clustering based adaptive multi-objective evolutionary algorithm & IEEE Transactions on Cybernetics & 2018\\
    \cite{tian2020coevolutionary} & \hyperref[CCMO]{CCMO} & Coevolutionary constrained multi-objective optimization framework & IEEE Transactions on Evolutionary Computation & 2020\\
    \cite{ming2021dual} & \hyperref[cDPEA]{cDPEA} & Constrained dual-population evolutionary algorithm & IEEE Transactions on Evolutionary Computation & 2021\\
    \cite{ge2018many} & \hyperref[CLIA]{CLIA} & Evolutionary algorithm with cascade clustering and reference point incremental learning & IEEE Transactions on Evolutionary Computation & 2018\\
    \cite{qiao2023evolutionaryconstrained} & \hyperref[CMEGL]{CMEGL} & Constrained evolutionary multitasking with global and local auxiliary tasks & IEEE/CAA Journal of Automatica Sinica & 2023\\
    \cite{zeng2023constrained} & \hyperref[CMODEFTR]{CMODEFTR} & Constrained multiobjective differential evolution based on the fusion of two rankings
     & Information Sciences & 2023\\
    \cite{jain2013evolutionary} & \hyperref[CMOEAD]{CMOEAD} & Constraint-MOEA/D & IEEE Transactions on Evolutionary Computation & 2013\\
    \cite{tian2021balancing} & \hyperref[CMOEAMS]{CMOEAMS} & Constrained multiobjective evolutionary algorithm with multiple stages & IEEE Transactions on Cybernetics & 2021\\
    \cite{zhang2018competitive} & \hyperref[CMOPSO]{CMOPSO} & Competitive mechanism based multi-objective particle swarm optimizer & Information Sciences & 2018\\
    \cite{he2022self} & \hyperref[CMOSMA]{CMOSMA} & Constrained multi-objective evolutionary algorithm with self-organizing map & Complex \& Intelligent Systems & 2022\\
    \cite{ming2021simple} & \hyperref[CTSEA]{CTSEA} & Constrained two-stage evolutionary algorithm & Knowledge-Based Systems & 2021\\
    \cite{jiao2020handling} & \hyperref[DCNSGAIII]{DCNSGAIII} & Dynamic constrained NSGA-III & IEEE Transactions on Cybernetics & 2020\\
    \cite{liu2019adapting} & \hyperref[DEAGNG]{DEAGNG} & Decomposition based evolutionary algorithm guided by growing neural gas & IEEE Transactions on Evolutionary Computation & 2019\\
    \cite{chen2017dmoea} & \hyperref[DMOEAeC]{DMOEAeC} & Decomposition-based multi-objective evolutionary algorithm with the e-constraint framework & IEEE Transactions on Evolutionary Computation & 2017\\
    \cite{zapotecas2011multi} & \hyperref[dMOPSO]{dMOPSO} & MOPSO based on decomposition & Proceedings of the Annual Conference on Genetic and Evolutionary Computation & 2011\\
    \cite{ming2024constrained} & \hyperref[DRLOSEMCMO]{DRLOSEMCMO} & EMCMO with deep reinforcement learning-assisted operator selection & IEEE/CAA Journal of Automatica Sinica & 2024\\
    \cite{yu2021dynamic} & \hyperref[DSPCMDE]{DSPCMDE} & Dynamic selection preference-assisted constrained multiobjective differential evolution
     & IEEE Transactions on Systems & 2021\\
    \cite{cai2014external} & \hyperref[EAGMOEAD]{EAGMOEAD} & External archive guided MOEA/D & IEEE Transactions on Evolutionary
    Computation & 2014\\
    \cite{yuan2015balancing} & \hyperref[EFRRR]{EFRRR} & Ensemble fitness ranking with a ranking restriction scheme & IEEE Transactions on Evolutionary Computation & 2015\\
    \cite{qiao2022evolutionary} & \hyperref[EMCMO]{EMCMO} & Evolutionary multitasking-based constrained multiobjective optimization & IEEE Transactions on Evolutionary Computation & 2022\\
    \cite{deb2003towards} & \hyperref[eMOEA]{eMOEA} & Epsilon multi-objective evolutionary algorithm & Proceedings of the International Conference on Evolutionary Multi-Criterion Optimization & 2003\\
    \cite{denysiuk2014clustering} & \hyperref[EMyOC]{EMyOC} & Evolutionary many-objective optimization algorithm with clustering-based selection & Proceedings of the International Conference on Parallel Problem Solving from Nature & 2014\\
    \cite{kukkonen2005gde3} & \hyperref[GDE3]{GDE3} & Generalized differential evolution 3 & Proceedings of the IEEE Congress on Evolutionary Computation & 2005\\
    \cite{tian2018guiding} & \hyperref[GFMMOEA]{GFMMOEA} & Generic front modeling based multi-objective evolutionary algorithm & IEEE Transactions on Cybernetics & 2018\\
    \cite{molina2009g} & \hyperref[gNSGAII]{gNSGAII} & g-dominance based NSGA-II & European Journal of Operational Research & 2009\\
    \cite{chen2019hyperplane} & \hyperref[hpaEA]{hpaEA} & Hyperplane assisted evolutionary algorithm & IEEE Transactions on Cybernetics & 2019\\
    \cite{bader2011hype} & \hyperref[HypE]{HypE} & Hypervolume estimation algorithm & Evolutionary Computation & 2011\\
    \cite{zitzler2004indicator} & \hyperref[IBEA]{IBEA} & Indicator-based evolutionary algorithm & Proceedings of the International Conference on Parallel Problem Solving from Nature & 2004\\
    \cite{yuan2021indicator} & \hyperref[ICMA]{ICMA} & Indicator-based constrained multi-objective algorithm & IEEE Transactions on Evolutionary Computation & 2021\\
    \cite{asafuddoula2014decomposition} & \hyperref[IDBEA]{IDBEA} & Improved decomposition-based evolutionary algorithm & IEEE Transactions on Evolutionary Computation & 2014\\
    \cite{qiao2023evolutionaryscalable} & \hyperref[IMTCMO]{IMTCMO} & Improved evolutionary multitasking-based CMOEA & IEEE Transactions on Evolutionary Computation & 2023\\
    \cite{tian2022local} & \hyperref[LMPFE]{LMPFE} & Evolutionary algorithm with local model based Pareto front estimation & IEEE Transactions on Systems & 2022\\
    \cite{he2016many} & \hyperref[MaOEACSS]{MaOEACSS} & Many-objective evolutionary algorithms based on coordinated selection
    strategy & IEEE Transactions on Evolutionary
    Computation & 2016\\
    \cite{cheng2015many} & \hyperref[MaOEADDFC]{MaOEADDFC} & Many-objective evolutionary algorithm based on directional diversity and
    favorable convergence & IEEE Transactions on Evolutionary Computation & 2015\\
    \cite{sun2018new} & \hyperref[MaOEAIT]{MaOEAIT} & Many-objective evolutionary algorithms based on an independent two-stage approach & IEEE Transactions on Evolutionary Computation & 2018\\
    \cite{zou2023multipopulation} & \hyperref[MCCMO]{MCCMO} & Multi-population coevolutionary constrained multi-objective optimization & IEEE Transactions on Evolutionary Computation & 2023\\
    \cite{jiao2022multiform} & \hyperref[MFOSPEA2]{MFOSPEA2} & Multiform optimization framework based on SPEA2 & IEEE Transactions on Cybernetics & 2022\\
    \cite{lin2015novel} & \hyperref[MMOPSO]{MMOPSO} & MOPSO with multiple search strategies & European Journal of Operational Research & 2015\\
    \cite{jiang2024mobca} & \hyperref[MOBCA]{MOBCA} & Multi-objective besiege and conquer algorithm & Biomimetics & 2024\\
    \cite{nebro2009mocell} & \hyperref[MOCell]{MOCell} & Cellular genetic algorithm & International Journal of Intelligent Systems & 2009\\
    \cite{igel2007covariance} & \hyperref[MOCMA]{MOCMA} & Multi-objective covariance matrix adaptation evolution strategy & Evolutionary computation & 2007\\
    \cite{zhang2007moea} & \hyperref[MOEAD]{MOEAD} & Multiobjective evolutionary algorithm based on decomposition & IEEE Transactions on Evolutionary Computation & 2007\\
    \cite{jiao2021two} & \hyperref[MOEAD2WA]{MOEAD2WA} & MOEA/D with two-type weight vector adjustments & Information Sciences & 2021\\
    \cite{qi2014moea} & \hyperref[MOEADAWA]{MOEADAWA} & MOEA/D with adaptive weight adjustment & Evolutionary Computation & 2014\\
    \cite{li2016biased} & \hyperref[MOEADCMA]{MOEADCMA} & MOEA/D with covariance matrix adaptation evolution strategy & IEEE Transactions on Cybernetics & 2016\\
    \cite{li2014evolutionary} & \hyperref[MOEADD]{MOEADD} & Many-objective evolutionary algorithm based on dominance and decomposition & IEEE Transactions on Evolutionary Computation & 2014\\
    \cite{zhu2020constrained} & \hyperref[MOEADDAE]{MOEADDAE} & MOEA/D with detect-and-escape strategy & IEEE Transactions on Evolutionary Computation & 2020\\
    \cite{takagi2019distribution} & \hyperref[MOEADDCWV]{MOEADDCWV} & MOEA/D with distribution control of weight vector set & Proceedings of the Bio-inspired Information and Communication Technologies & 2019\\
    \cite{li2008multiobjective} & \hyperref[MOEADDE]{MOEADDE} & MOEA/D based on differential evolution & IEEE Transactions on Evolutionary Computation & 2008\\
    \cite{zhang2009performance} & \hyperref[MOEADDRA]{MOEADDRA} & MOEA/D with dynamical resource allocation & Proceedings of the IEEE Congress on Evolutionary Computation & 2009\\
    \cite{yuan2015balancing} & \hyperref[MOEADDU]{MOEADDU} & MOEA/D with a distance based updating strategy & IEEE Transactions on Evolutionary Computation & 2015\\
    \cite{sun2020adaptive} & \hyperref[MOEADDYTS]{MOEADDYTS} & MOEA/D with dynamic Thompson sampling & Proceedings of the International Conference on
    Parallel Problem Solving from Nature & 2020\\
    \cite{li2013adaptive} & \hyperref[MOEADFRRMAB]{MOEADFRRMAB} & MOEA/D with fitness-rate-rank-based multiarmed bandit & IEEE Transactions on Evolutionary Computation & 2013\\
    \cite{liu2013decomposition} & \hyperref[MOEADM2M]{MOEADM2M} & MOEA/D based on MOP to MOP & IEEE Transactions on Evolutionary Computation & 2013\\
    \cite{gee2014online} & \hyperref[MOEADMRDL]{MOEADMRDL} & MOEA/D with maximum relative diversity loss & IEEE Transactions on Evolutionary Computation & 2014\\
    \cite{wang2016decomposition} & \hyperref[MOEADPaS]{MOEADPaS} & MOEA/D with Pareto adaptive scalarizing approximation & IEEE Transactions on Evolutionary Computation & 2016\\
    \cite{li2013stable} & \hyperref[MOEADSTM]{MOEADSTM} & MOEA/D with stable matching & IEEE Transactions on Evolutionary Computation & 2013\\
    \cite{de2022decomposition} & \hyperref[MOEADUR]{MOEADUR} & MOEA/D with update when required & Swarm and Evolutionary Computation & 2022\\
    \cite{farias2019many} & \hyperref[MOEADURAW]{MOEADURAW} & MOEA/D with uniform randomly adaptive weights & In Proceedings of the IEEE International Conference on Systems & 2019\\
    \cite{takagi2021weight} & \hyperref[MOEADVOV]{MOEADVOV} & MOEA/D with virtual objective vectors & Proceedings of the IEEE Congress on Evolutionary Computation & 2021\\
    \cite{tian2016multi} & \hyperref[MOEAIGDNS]{MOEAIGDNS} & Multi-objective evolutionary algorithm based on an enhanced IGD & Proceedings of the IEEE Congress on Evolutionary Computation & 2016\\
    \cite{denysiuk2015moea} & \hyperref[MOEAPC]{MOEAPC} & Multiobjective evolutionary algorithm based on polar coordinates & Proceedings of the International Conference on Evolutionary Multi-Criterion Optimization & 2015\\
    \cite{hernandez2015improved} & \hyperref[MOMBIII]{MOMBIII} & Many objective metaheuristic based on the R2 indicator II & Proceedings of the Annual Conference on Genetic and Evolutionary Computation & 2015\\
    \cite{coello2002mopso} & \hyperref[MOPSO]{MOPSO} & Multi-objective particle swarm optimization & Proceedings of the IEEE Congress on Evolutionary Computation & 2002\\
    \cite{raquel2005effective} & \hyperref[MOPSOCD]{MOPSOCD} & MOPSO with crowding distance & Proceedings of the Annual Conference on Genetic and Evolutionary Computation & 2005\\
    \cite{knowles2000m} & \hyperref[MPAES]{MPAES} & Memetic algorithm with Pareto archived evolution strategy & Proceedings of the IEEE Congress on Evolutionary Computation & 2000\\
    \cite{dai2015new} & \hyperref[MPSOD]{MPSOD} & Multi-objective particle swarm optimization algorithm based on decomposition & Information Sciences & 2015\\
    \cite{zhang2023design} & \hyperref[MSCEA]{MSCEA} & Multi-stage constrained multi-objective evolutionary algorithm & Information Sciences & 2023\\
    \cite{ma2021multi} & \hyperref[MSCMO]{MSCMO} & Multi-stage constrained multi-objective evolutionary algorithm & Information Sciences & 2021\\
    \cite{tian2019multistage} & \hyperref[MSEA]{MSEA} & Multi-stage multi-objective evolutionary algorithm & IEEE Transactions on Systems & 2019\\
    \cite{hughes2007msops} & \hyperref[MSOPSII]{MSOPSII} & Multiple single objective Pareto sampling II & Proceedings of the IEEE Congress on Evolutionary Computation & 2007\\
    \cite{qiao2022dynamic} & \hyperref[MTCMO]{MTCMO} & Multitasking constrained multi-objective optimization & IEEE Transactions on Evolutionary Computation & 2022\\
    \cite{tseng2009multiple} & \hyperref[MTS]{MTS} & Multiple trajectory search & Proceedings of the IEEE Congress on Evolutionary Computation & 2009\\
    \cite{denysiuk2013many} & \hyperref[MyODEMR]{MyODEMR} & Many-objective differential evolution with mutation restriction & Proceedings of the Annual Conference on Genetic and Evolutionary Computation & 2013\\
    \cite{lin2016particle} & \hyperref[NMPSO]{NMPSO} & Novel multi-objective particle swarm optimization & IEEE Transactions on Evolutionary Computation & 2016\\
    \cite{gong2008multiobjective} & \hyperref[NNIA]{NNIA} & Nondominated neighbor immune algorithm & Evolutionary Computation & 2008\\
    \cite{hua2024adaptive} & \hyperref[NRVMOEA]{NRVMOEA} & Adaptive normal reference vector-based multi- and many-objective evolutionary algorithm
     & Complex \& Intelligent Systems & 2024\\
    \cite{mendes2023non} & \hyperref[NSBiDiCo]{NSBiDiCo} & Non-dominated sorting bidirectional differential coevolution algorithm & Proceedings of the IEEE International Conference on Systems & 2023\\
    \cite{deb2002fast} & \hyperref[NSGAII]{NSGAII} & Nondominated sorting genetic algorithm II & IEEE Transactions on Evolutionary Computation & 2002\\
    \cite{pan2021adaptive} & \hyperref[NSGAIIARSBX]{NSGAIIARSBX} & NSGA-II with adaptive rotation based simulated binary crossover & Swarm and Evolutionary Computation & 2021\\
    \cite{deb2013evolutionary} & \hyperref[NSGAIII]{NSGAIII} & Nondominated sorting genetic algorithm III & IEEE Transactions on Evolutionary Computation & 2013\\
    \cite{chen2014new} & \hyperref[NSLS]{NSLS} & Multiobjective optimization framework based on nondominated sorting and local search & IEEE Transactions on Evolutionary Computation & 2014\\
    \cite{liu2017many} & \hyperref[onebyoneEA]{onebyoneEA} & Many-objective evolutionary algorithm using a one-by-one selection strategy & IEEE Transactions on Cybernetics & 2017\\
    \cite{li2021estimation} & \hyperref[PeEA]{PeEA} & Pareto front shape estimation based evolutionary algorithm & Information Sciences & 2021\\
    \cite{corne2001pesa} & \hyperref[PESAII]{PESAII} & Pareto envelope-based selection algorithm II & Proceedings of the Annual Conference on Genetic and Evolutionary Computation & 2001\\
    \cite{wang2012preference} & \hyperref[PICEAg]{PICEAg} & Preference-inspired coevolutionary algorithm with goals & IEEE Transactions on Evolutionary Computation & 2012\\
    \cite{fan2019push} & \hyperref[PPS]{PPS} & Push and pull search algorithm & Swarm and Evolutionary Computation & 2019\\
    \cite{yuan2020investigating} & \hyperref[PREA]{PREA} & Promising-region based EMO algorithm & IEEE Transactions on Evolutionary Computation & 2020\\
    \cite{zhang2008rm} & \hyperref[RMMEDA]{RMMEDA} & Regularity model-based multiobjective estimation of distribution algorithm
     & IEEE Transactions on Evolutionary Computation & 2008\\
    \cite{said2010r} & \hyperref[rNSGAII]{rNSGAII} & r-dominance based NSGA-II & IEEE Transactions on Evolutionary Computation & 2010\\
    \cite{elarbi2017new} & \hyperref[RPDNSGAII]{RPDNSGAII} & Reference point dominance-based NSGA-II & IEEE Transactions on Systems & 2017\\
    \cite{he2017radial} & \hyperref[RSEA]{RSEA} & Radial space division based evolutionary algorithm & Applied Soft Computing & 2017\\
    \cite{cheng2016reference} & \hyperref[RVEA]{RVEA} & Reference vector guided evolutionary algorithm & IEEE Transactions on Evolutionary Computation & 2016\\
    \cite{cheng2016reference} & \hyperref[RVEAa]{RVEAa} & RVEA embedded with the reference vector regeneration strategy & IEEE Transactions on Evolutionary Computation & 2016\\
    \cite{liu2020adaptive} & \hyperref[RVEAiGNG]{RVEAiGNG} & RVEA based on improved growing neural gas & IEEE Transactions on Cybernetics & 2020\\
    \cite{zhang2016self} & \hyperref[SMEA]{SMEA} & Self-organizing multiobjective evolutionary algorithm & IEEE Transactions on Evolutionary Computation & 2016\\
    \cite{nebro2009smpso} & \hyperref[SMPSO]{SMPSO} & Speed-constrained multi-objective particle swarm optimization & Proceedings of the IEEE Symposium on Computational Intelligence in Multi-Criteria Decision-Making & 2009\\
    \cite{emmerich2005emo} & \hyperref[SMSEMOA]{SMSEMOA} & S metric selection based evolutionary multiobjective optimization algorithm & Proceedings of the International Conference on Evolutionary Multi-Criterion Optimization & 2005\\
    \cite{zitzler2001spea2} & \hyperref[SPEA2]{SPEA2} & Strength Pareto evolutionary algorithm 2 & Proceedings of the Conference on Evolutionary Methods for Design & 2001\\
    \cite{jiang2017strength} & \hyperref[SPEAR]{SPEAR} & Strength Pareto evolutionary algorithm based on reference direction & IEEE Transactions on Evolutionary Computation & 2017\\
    \cite{liu2023subspace} & \hyperref[SSCEA]{SSCEA} & Subspace segmentation based co-evolutionary algorithm & Swarm and Evolutionary Computation & 2023\\
    \cite{yuan2015new} & \hyperref[tDEA]{tDEA} & theta-dominance based evolutionary algorithm & IEEE Transactions on Evolutionary Computation & 2015\\
    \cite{ming2023constraint} & \hyperref[tDEACPBI]{tDEACPBI} & Theta-dominance based evolutionary algorithm with CPBI & IEEE Transactions on Systems & 2023\\
    \cite{liu2019handling} & \hyperref[ToP]{ToP} & Two-phase framework with NSGA-II & IEEE Transactions on Evolutionary Computation & 2019\\
    \cite{ming2022two} & \hyperref[TSNSGAII]{TSNSGAII} & Two stage NSGA-II & IEEE Transactions on Systems & 2022\\
    \cite{dong2022two} & \hyperref[TSTI]{TSTI} & Two-stage evolutionary algorithm with three indicators & Expert Systems with Applications & 2022\\
    \cite{wang2014two_arch2} & \hyperref[TwoArch2]{Two\_Arch2} & Two-archive algorithm 2 & IEEE Transactions on Evolutionary Computation & 2014\\
    \cite{liang2022utilizing} & \hyperref[URCMO]{URCMO} & Utilizing the relationship between constrained and unconstrained Pareto fronts for constrained multi-objective optimization
     & IEEE Transactions on Cybernetics & 2022\\
    \cite{xiang2016vector} & \hyperref[VaEA]{VaEA} & Vector angle based evolutionary algorithm & IEEE Transactions on Evolutionary Computation & 2016\\
    \cite{zhang2016weight} & \hyperref[WVMOEAP]{WVMOEAP} & Weight vector based multi-objective optimization algorithm with preference & Acta Electronica Sinica (Chinese) & 2016\\
    \bottomrule
\end{longtable}
}

\newread\file
\openin\file=merged2.txt

\clearpage
\subsection{Additional results not included in the main paper}
\begin{figure}[!h]
    \centering
    \begin{subfigure}[b]{0.48\linewidth}
        \centering
        \includegraphics[width=\linewidth]{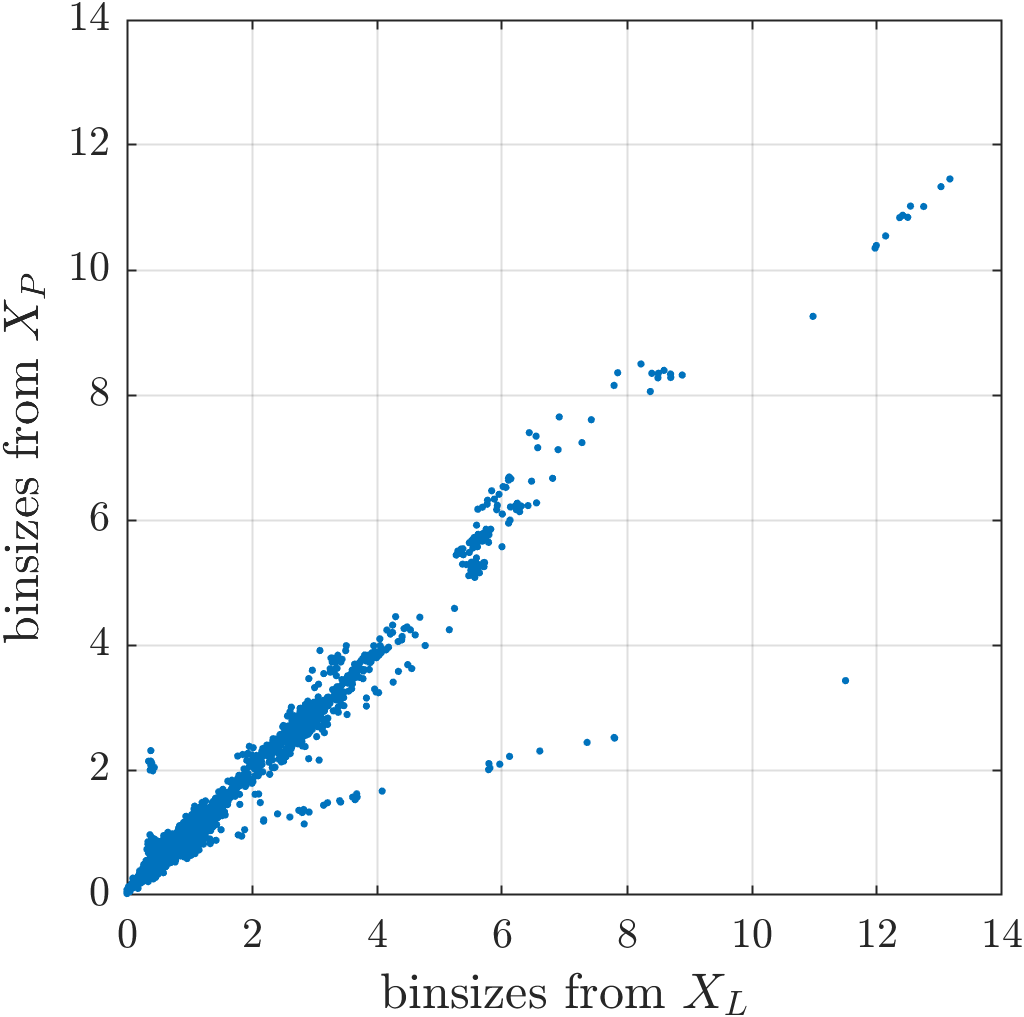}
    \end{subfigure}\hfill
    \begin{subfigure}[b]{0.48\linewidth}
        \centering
        \includegraphics[width=\linewidth]{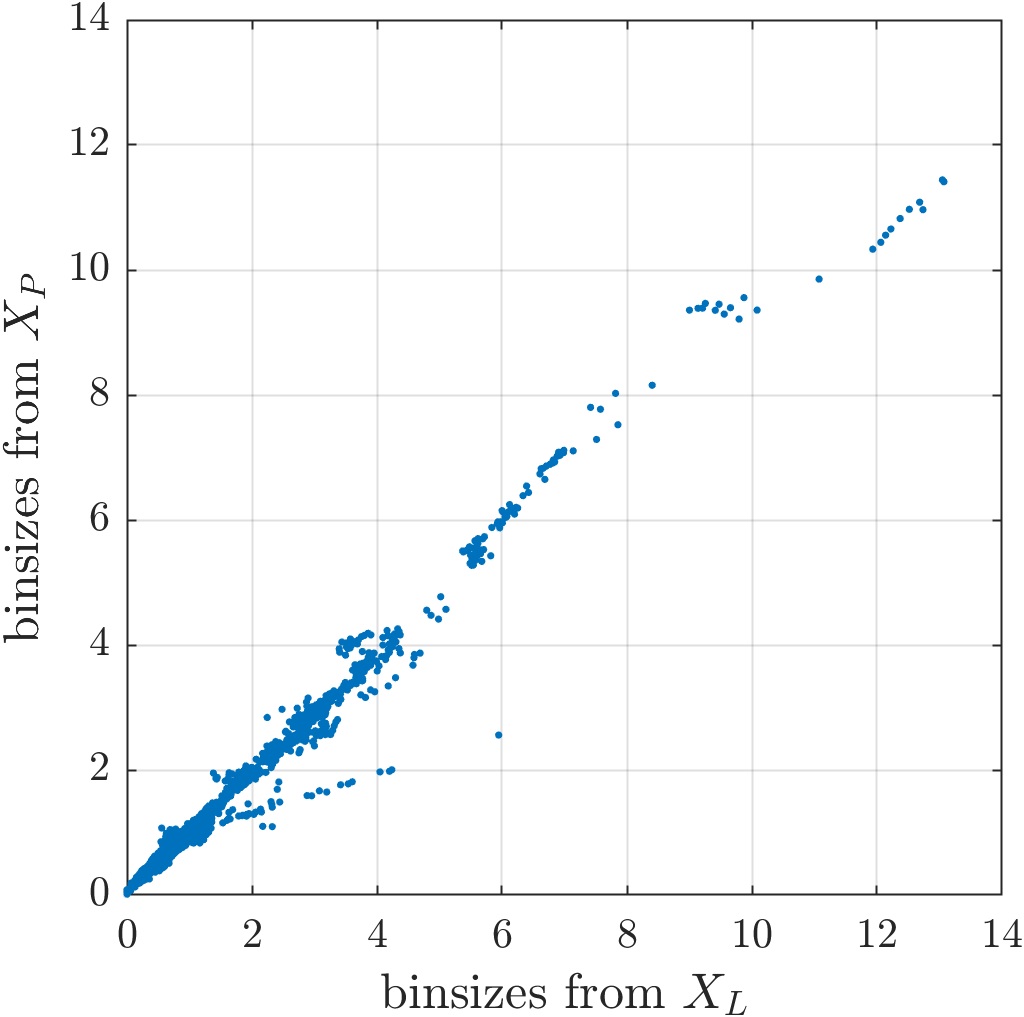}
    \end{subfigure}
    \caption{Relation between $X_L$ and $X_P$ relative binsizes are roughly the same, which justifies using $X_P$ for the computation of BIAS, chi2, etc., $d=2$ on the left and $d=10$ on the right.}
\end{figure}

\begin{figure}[!h]
    \centering
    \begin{subfigure}[b]{0.48\linewidth}
        \centering
        \includegraphics[width=1\linewidth,trim=0mm 0mm 130mm 0mm,clip]{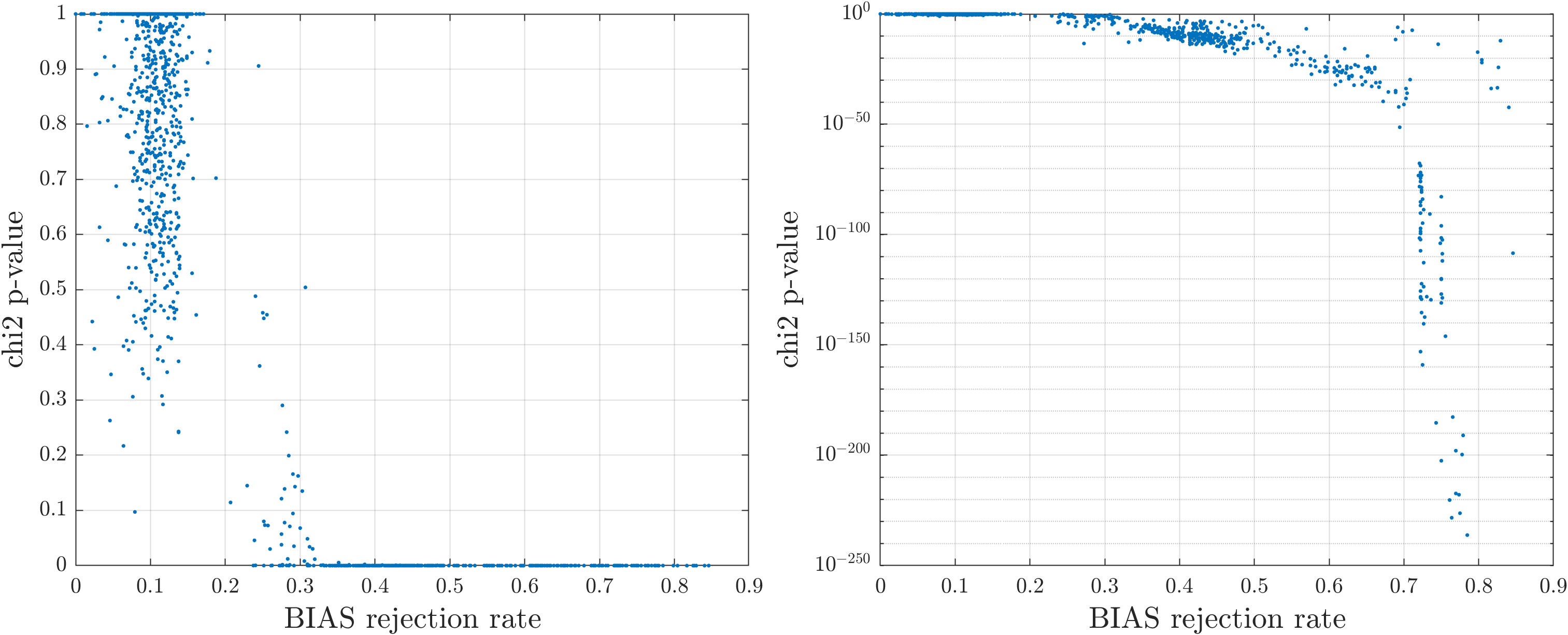}
    \end{subfigure}\hfill
    \begin{subfigure}[b]{0.48\linewidth}
        \centering
        \includegraphics[width=1\linewidth,trim=0mm 0mm 130mm 0mm,clip]{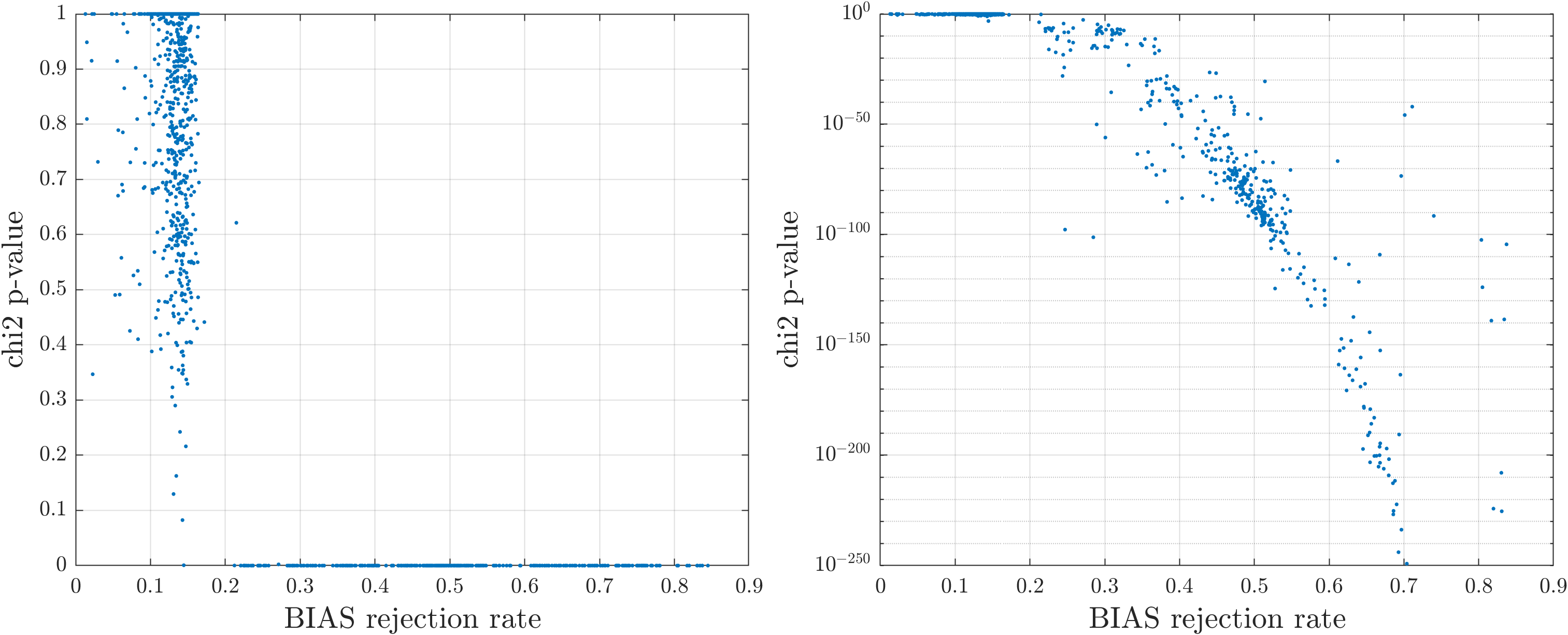}
    \end{subfigure}
    \caption{Relashionship between average BIAS rejection rate and geometric average of chi2 p value, $d=2$ on the left and $d=10$ on the right.}
\end{figure}

\begin{figure}[!h]
    \centering
    \includegraphics[width=0.49\linewidth]{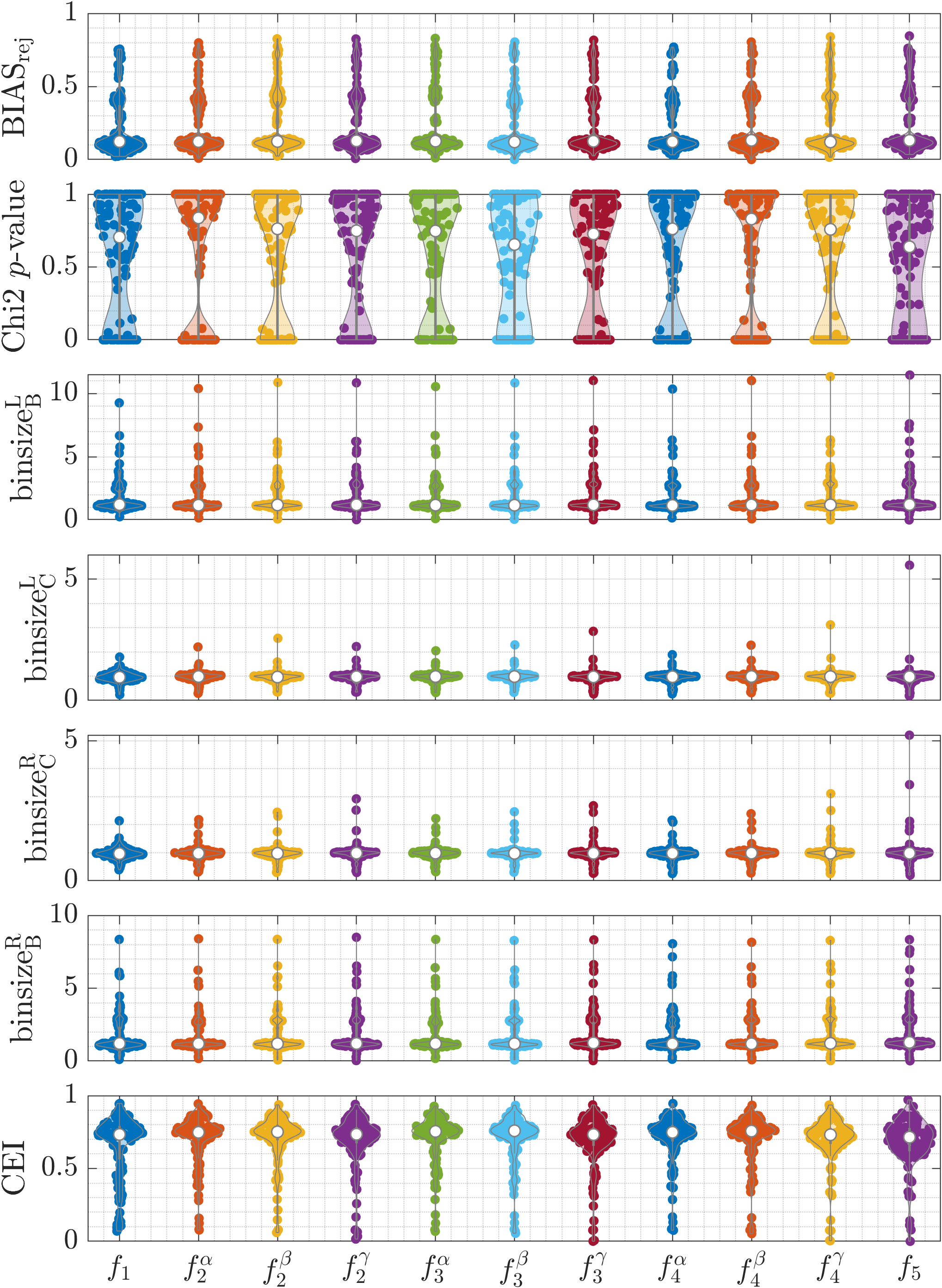}
    \includegraphics[width=0.49\linewidth]{fig_violin_d10.png}
    \caption{Distribution of values of the tracked parameters used for detection of structural bias, $d=2$ (left), $d=10$ (right).} 
\end{figure}

\clearpage
\subsection{Detailed results per algorithm}
This section contains results for all algorithms listed in Section~\ref{sect:algorithms} for experiments in dimensionalities $d=2$ and $d=10$, described in the main paper. The meaning of the columns is as follows, with the colour scheme is detailed in Table~\ref{tab:colorscheme}: 
\begin{itemize}
    \item \textbf{$\text{BIAS}_\text{rej}$}: Proportion of rejections from the statistical test of uniformity from the BIAS toolbox. Computed on $X_P$.
    \item \textbf{Chi2 p-value}:  Geometric average of the $p$-values from the chi-squared test, scaled by a factor of $e$. Computed on $X_P$.
    \item \textbf{$\text{binsize}_B^L$}: Relative height of the left bound bin from the aggregate histogram of $X_P$.
    \item \textbf{$\text{binsize}_C^L$}: Relative height of the left central bin from the aggregate histogram of $X_P$.
    \item \textbf{$\text{binsize}_C^R$}: Relative height of the right central bin from the aggregate histogram of $X_P$.
    \item \textbf{$\text{binsize}_B^R$}: Relative height of the right bound bin from the aggregate histogram of $X_P$.
    \item \textbf{CEI}: Clark-Evans Index, computed on $X_L$.
\end{itemize}

\begin{table}[!h]
    \centering
    \caption{Values for color scheme (Chi2 $p$-value is log scaled)}\label{tab:colours}
    \begin{tabular}{cccccc}
    $\text{BIAS}_\text{rej}$ &\cellcolor[HTML]{fde725}$\leq$0.1& \cellcolor[HTML]{5ec962}0.2& \cellcolor[HTML]{21918c}\color{white}0.3&\cellcolor[HTML]{3b528b}\color{white}0.5&\cellcolor[HTML]{440154}\color{white}1\\
    Chi2 $p$-value &\cellcolor[HTML]{fde725}$\geq$0.1& \cellcolor[HTML]{5ec962}0.05& \cellcolor[HTML]{21918c}\color{white}0.01&\cellcolor[HTML]{3b528b}\color{white}1E-20&\cellcolor[HTML]{440154}\color{white}$\leq$1E-50\\
    $\text{binsize}_B^L$ &\cellcolor[HTML]{fde725}$\leq$1& \cellcolor[HTML]{5ec962}1.2& \cellcolor[HTML]{21918c}\color{white}2&\cellcolor[HTML]{3b528b}\color{white}4&\cellcolor[HTML]{440154}\color{white}$\geq$10\\
    $\text{binsize}_C^L$ &\cellcolor[HTML]{fde725}$\leq$1& \cellcolor[HTML]{5ec962}1.2& \cellcolor[HTML]{21918c}\color{white}2&\cellcolor[HTML]{3b528b}\color{white}4&\cellcolor[HTML]{440154}\color{white}$\geq$10\\
    $\text{binsize}_C^R$ &\cellcolor[HTML]{fde725}$\leq$1& \cellcolor[HTML]{5ec962}1.2& \cellcolor[HTML]{21918c}\color{white}2&\cellcolor[HTML]{3b528b}\color{white}4&\cellcolor[HTML]{440154}\color{white}$\geq$10\\
    $\text{binsize}_B^R$ &\cellcolor[HTML]{fde725}$\leq$1& \cellcolor[HTML]{5ec962}1.2& \cellcolor[HTML]{21918c}\color{white}2&\cellcolor[HTML]{3b528b}\color{white}4&\cellcolor[HTML]{440154}\color{white}$\geq$10\\
    CEI &\cellcolor[HTML]{fde725}1 &\cellcolor[HTML]{5ec962}0.8& \cellcolor[HTML]{21918c}\color{white}0.6&\cellcolor[HTML]{3b528b}\color{white}0.4&\cellcolor[HTML]{440154}\color{white}$\leq$0.1\\
    \end{tabular}
    \label{tab:colorscheme}
\end{table}

\clearpage

\clearpage
\foreach \i in {1,...,120} {
    \read\file to\fileline
    \fileline
    \\[1cm]
    \noindent
    \pgfplotstabletypeset[
        col sep=comma,
        string type,
        header = false,
        columns/0/.style = {column type=l|},
        every nth row={30[+1]}{before row=\midrule},
        every head row/.style={output empty row,after row=\bottomrule},%
        every last row/.style={
            after row=\bottomrule
        }
    ]{tab\i_d2.csv}
    \\[1cm]
    \noindent
    \pgfplotstabletypeset[
        col sep=comma,
        string type,
        header = false,
        columns/0/.style = {column type=l|},
        every nth row={30[+1]}{before row=\midrule},
        every head row/.style={output empty row,after row=\bottomrule},%
        every last row/.style={
            after row=\bottomrule
        }
    ]{tab\i_d10.csv}
    \clearpage 
}
\closein\file

\putbib[biblio_appendix]
\end{bibunit}
\end{document}